\documentclass[review,times]{elsarticle}
\usepackage[utf8]{inputenc}
\usepackage[T1]{fontenc}
\usepackage{multirow}
\usepackage{bm}
\usepackage{amssymb}
\usepackage{mathtools}
\usepackage{pdflscape}
\usepackage{threeparttable}
\usepackage{rotating}
\usepackage{subcaption}
\usepackage{url}

\begin{document}

\title{ A few filters are enough: Convolutional Neural Network for P300 Detection}

\author[mymainaddress]{Montserrat Alvarado-González\fnref{fn1}}
\cortext[mycorrespondingauthor]{Corresponding author} \ead{amontserrat@gmail.com} 

\author[mysecondaryaddress]{Gibran Fuentes-Pineda\fnref{fn1}}
\author[mymainaddress]{Jorge Cervantes-Ojeda}  
\fntext[fn1]{These authors contributed equally to this work.}
\address[mymainaddress]{Department of Applied Mathematics and Systems, Universidad Autónoma Metropolitana, 05348, Mexico City, Mexico}

\address[mysecondaryaddress]{Department of Computer Science, Instituto de Investigaciones en Matemáticas Aplicadas y en Sistemas, Universidad Nacional Autónoma de México, 04510, Mexico City, Mexico}

\begin{abstract}
Over the past decade convolutional neural networks (CNNs) have become the driving force of an ever-increasing set of applications, achieving state-of-the-art performance. Modern CNN architectures are often composed of many convolutional and some fully connected layers, and have thousands or millions of parameters. CNNs have shown to be effective in the detection of Event-Related Potentials from electroencephalogram (EEG) signals, notably the P300 component which is frequently employed in Brain-Computer Interfaces (BCIs). However, for this task, the increase in detection rates compared to approaches based on human-engineered features has not been as impressive as in other areas and might not justify such a large number of parameters. In this paper, we study the performance of existing CNN architectures with diverse complexities for single-trial within-subject and cross-subject P300 detection on four different datasets. We also proposed SepConv1D, a very simple CNN architecture consisting of a single depthwise separable 1D convolutional layer followed by a fully connected Sigmoid classification neuron. We found that with as few as four filters in its convolutional layer and an overall small number of parameters, SepConv1D obtained competitive performances in the four datasets. We believe these results may represent an important step towards building simpler, cheaper, faster, and more portable BCIs.

\end{abstract}
\begin{keyword}Convolutional Neural Network, P300, single-trial, within-subject, cross-subject
\end{keyword}
\maketitle

\section{Introduction}
A Brain-Computer Interface (BCI) is a system composed of software and hardware that builds a channel of communication between the subject and the computer, using only the subject's brain signals~\citep{Wolpaw2000}. The general process followed by a BCI can be divided into: subject's stimulation, acquisition of brain activity, signal preprocessing, feature extraction, classification, and translation of the output's classification into instructions to control an application or a device. One of the brain signals, captured with the electroencephalogram (EEG), of interest in the community for controlling BCIs is the Event-Related Potential (ERP). In particular, the late P300 component has a stable temporal relationship with respect to the stimulation event, an interesting characteristic to control BCIs. For instance, it is related to cognitive and attention processes and it is independent of the type of stimulation presented to the subject. The P300 is associated with the oddball paradigm, which consists in presenting a series of frequent stimuli interrupted by infrequent stimuli \citep{Donchin2000}. Thus, every time the infrequent stimulus is detected by the subject, the brain unconsciously generates a positive peak, approximately 300 ms afterwards.

Given that a BCI based on ERP signals is highly subject-specific, it requires two phases to be used: \textit{i)} an offline training phase to calibrate the system for each subject and \textit{ii)} an online phase to actually let the subject control the BCI. Typically, the subject has to repeat several times the oddball paradigm to increase the ERP's signal-to-noise ratio \citep{Niedermeyer2005}. Since the stimulation process can become unacceptably slow and tiring for the subject, much of the effort in the BCI development is to stimulate the subject as few times as possible, preferably only once (i.e., by a single-trial), while still achieving an adequate detection of the P300 component. More recently, some works have attempted to eliminate the calibration stage for each subject by using instead the information acquired previously for other subjects~\citep{Lawhern2018}. The P300 detection based on the information retrieved during the calibration stage by a single subject is known as within-subject classification, whereas the detection based on the information retrieved by other subjects is known as cross-subject classification.  

In order to detect the P300 under these conditions, different domains have been used to represent ERP's features, including frequency \cite{Kaur2013}, time-frequency \cite{Atum2010}, space-time \cite{Farwell1988}, and shape \cite{Alvarado2016b}. Prior to classification, feature selection is commonly applied to: \textit{i)} reduce redundancy, \textit{ii)} choose the ones related to the mental states targeted by the BCI, \textit {iii)}  generate fewer parameters to be optimized by the classifier, and \textit{iv)} produce faster predictions for a new sample. Among the most prominent feature selection approaches used for P300 detection are the embedded methods (e.g., Stepwise Linear Discriminant Analysis \citep{Krusienski2006}) and the wrapper methods (e.g., Genetic Algorithms \citep{Atum2010}). On the other hand, the approaches more commonly used for classification have been Linear Discriminant Analysis \citep{Bostanov2004}, Support Vector Machines \citep{Kaper2004}, Feed Forward Neural Networks \citep{Cechovic2013, Abdulhay2017}, and adaptive classifiers \citep{Woehrle2015,Zeyl2016}. 

Recently, some methods merge feature extraction, feature selection, and classification by using matrix classifiers (e.g.~\citep{Mayaud2016}) or Deep Learning (e.g. ~\citep{Schirrmeister2017, Liu2018, Lawhern2018}). In particular, the latter has gained a lot of interest since it has demonstrated to be very effective in fields such as Computer Vision~\citep{He2016} and Speech Recognition~\citep{Abdel-Hamid2014}, not only to replace human-engineered features but also to increase classification rates. Some characteristics of these methods are their depth, the use of a large number of parameters, and the need of huge amounts of data to train. These characteristics may be a disadvantage for P300 detection, mainly due to the limited training data available \citep{Lotte2015b, Lotte2018}. 
 
Although CNN architectures have been effective for single-trial within-subject and cross-subject P300 detection from EEG signals, the increase in detection rates compared to approaches based on human-engineered features has not been as impressive as in other applications~\citep{Roy2019} and might not justify such a large number of parameters. In this paper, we study the performance of state-of-the-art CNN architectures with diverse complexities for single-trial within-subject and cross-subject P300 detection on four different datasets. We also propose SepConv1D, a very simple CNN architecture consisting of a single depthwise separable 1D convolutional layer followed by a fully connected Sigmoid classification neuron. We compare the state-of-the-art architectures to SepConv1D and a simple Fully-Connected Neural Network (FCNN) with a single hidden layer with only two neurons, both in terms of detection performance and complexity.

The remainder of this paper is organized as follows. In Section \ref{sec:State-of-the-art}, we review the state-of-the-art CNN architectures for P300 detection.  In Section \ref{sec:Methods}, we describe in detail SepConv1D and the Fully-Connected Neural Network. In Section \ref{sec:Experimental_Design}, we present the experimental design and the datasets. Section \ref{sec:Results} shows the results of the experimental evaluation and discusses the performances and complexities of the analyzed architectures. Finally, in Section \ref{sec:Conclusions} we provide some concluding remarks.

\section{State-of-the-art CNN architectures to detect P300} \label{sec:State-of-the-art}
Convolutional Neural Networks (CNN) have become the state-of-the-art for single-trial P300 detection from EEG signals. Recently, many different CNN architectures have been proposed for this task, achieving high detection performances. This section gives a concise presentation of these architectures, specifying the minor modifications we made in some of them for comparison purposes. Details about the activation functions can be found in~\ref{sec:Appendix}.

\paragraph{CNN-1} Cecotti et al.~\citep{Cecotti2011} proposed a 4-layer architecture, named CNN-1. To the best of our knowledge, this was the first CNN for P300 detection. The first layer of CNN-1 is a convolution in the space domain, which learns combinations of the input channels. The second layer is a 1D-convolution over time which subsamples and filters the feature maps obtained by the first convolutional layer.
Both convolutional layers have a Scaled Hyperbolic Tangent activation function. The output of the second convolutional layer is flattened and fed into a fully-connected layer with 100 neurons. 
Finally, the classification layer is composed of two Sigmoid neurons. Table \ref{tab:CNN-1_Arq} shows further details about the CNN-1 architecture. The authors also presented CNN-3, a variation of CNN-1 that has only one filter in the first convolutional layer, in contrast to the 10 filters of CNN-1. See Table \ref{tab:CNN-3_Arq} to compare the differences in the number of parameters and the outputs of the layers.
To evaluate the performance of these architectures in the current research context, we slightly modified these two architectures as follows: \textit{i)} we changed the classification layer from two neurons with a Sigmoid activation function to two neurons with a Softmax activation function, and \textit{ii)} we changed the loss function from Mean Squared Error to Categorical Cross Entropy. We named the modified CNN-1 and CNN-3 as UCNN-1 and UCNN-3 respectively.

\paragraph{CNN-R}  
Manor and Geva \citep{Manor2015} proposed a more complex architecture, containing two convolutional-max-pooling blocks, a convolutional layer, two fully connected layers, and a Softmax classification layer. The first convolutional block performs a spatial convolution. The weights and biases of this layer are regularized with a spatiotemporal penalty that reduces overfitting. The second convolutional block finds temporal patterns that represent the change in amplitude of the spatial maps learned in the first block. CNN-R uses a ReLU activation function in all its hidden layers and applies Dropout after each fully-connected layer. Table \ref{tab:CNNR_Arq} depicts the detailed CNN-R's architecture. 

\paragraph{DeepConvNet} Schirrmeister et al. \citep{Schirrmeister2017} proposed an architecture consisting of four convolutional blocks followed by a Softmax classification layer. The first convolutional block is composed of two layers: in the first layer each filter performs a convolution over time and in the second each filter performs a spatial filtering. The first convolutional block is followed by three standard convolutional-max-pooling blocks. DeepConvNet uses ELU as activation function in all its hidden layers and Dropout is applied after each convolutional block. Table \ref{tab:DeepConvNet_Arq} shows further details about the architecture. 

\paragraph{ShallowConvNet}  In addition to DeepConvNet, Schirrmeister et al.~\citep{Schirrmeister2017} presented an architecture originally designed to decode band-power features. The first two layers of this architecture perform a temporal and spatial convolution. They are followed by a squaring nonlinearity, an average pooling operation, a logarithmic activation function, and a Softmax classification layer. In ShallowConvNet, Dropout is applied before the Softmax classification layer. See Table \ref{tab:ShallowConvNet_Arq} for more details about the architecture.

\paragraph{BN$^3$} 
 Liu et al. \citep{Liu2018} proposed a six-layer CNN architecture that combines Batch Normalization \citep{Ioffe2015} and Dropout \citep{Srivastava2014} techniques.
 First, it applies a 1D convolutional layer for spatial feature extraction, followed by a 1D convolutional and a subsampling layer for temporal feature extraction. The two convolutional layers are followed by a Batch Normalization, two fully connected layers with 128 neurons and a classification layer with a single Sigmoid neuron. The ReLU activation function is used in the second convolutional layer and the Hyperbolic Tangent activation function is used in the two fully connected layers. Dropout is applied after each fully-connected layer. Table \ref{tab:BN3_Arq} depicts details about the architecture.

\paragraph{EEGNet} 
Lawhern et al.  \citep{Lawhern2018} proposed a compact CNN architecture  consisting of two convolutional blocks followed by a Softmax classification layer. The first convolutional block decomposes the EEG signal at different band-pass frequencies and reduces the number of trainable parameters by a depthwise convolution. The architecture applies Batch Normalization \citep{Ioffe2015} along the feature map dimension before employing an ELU activation function. The second convolutional block uses a separable convolution, which is a depthwise convolution followed by point-wise convolutions, to reduce the number of parameters and to decouple the relationship between feature maps. EEGNet applies Average Pooling and Dropout after each convolutional block. See Table \ref{tab:EEGNet_Arq} for more details about the architecture. 

\paragraph{OCLNN} 
Shan et al.~\citep{Shan2018} proposed a one-convolutional-layer architecture, which is the simplest CNN-based architecture for P300 detection in terms of number of layers. It consists of a 1D convolutional layer with a ReLU activation function, followed by a Softmax classification layer. The convolutional layer divides the temporal signals from the input channels into 15 parts and performs a convolution operation for temporal and spatial feature extraction. Dropout is applied before the Softmax classification layer. See Table \ref{tab:OCLNN_Arq} for more details about the architecture.  

\section{Methods} \label{sec:Methods}
As previously explained, we would like to provide an alternative to complex architectures for P300 detection. To that end, we now describe a simple CNN-based architecture, the SepConv1D. Additionally, we describe a one-layer Fully-Connected Neural Network. 

\subsection{EEG signals}
The EEG signals are recorded by $C$ channels and discretized by $S$ temporal signal samples. From now on,  $S=T \times F$, where $T$ corresponds to the time period between the moment posterior to the stimulus and $T$, and $F$ denotes the signal sampling frequency. The filtered signals are represented by a matrix $\mathbf{X}$ of size $C \times T$, where each vector $\mathbf{x}_{c}= [x_{1},\dots,x_{T}]$  represents the signals recorded by a single EEG channel $c\in \{1,\dots,C\}$.

\subsection{SepConv1D}

\begin{figure}
\begin{centering}
\includegraphics[scale=0.3]{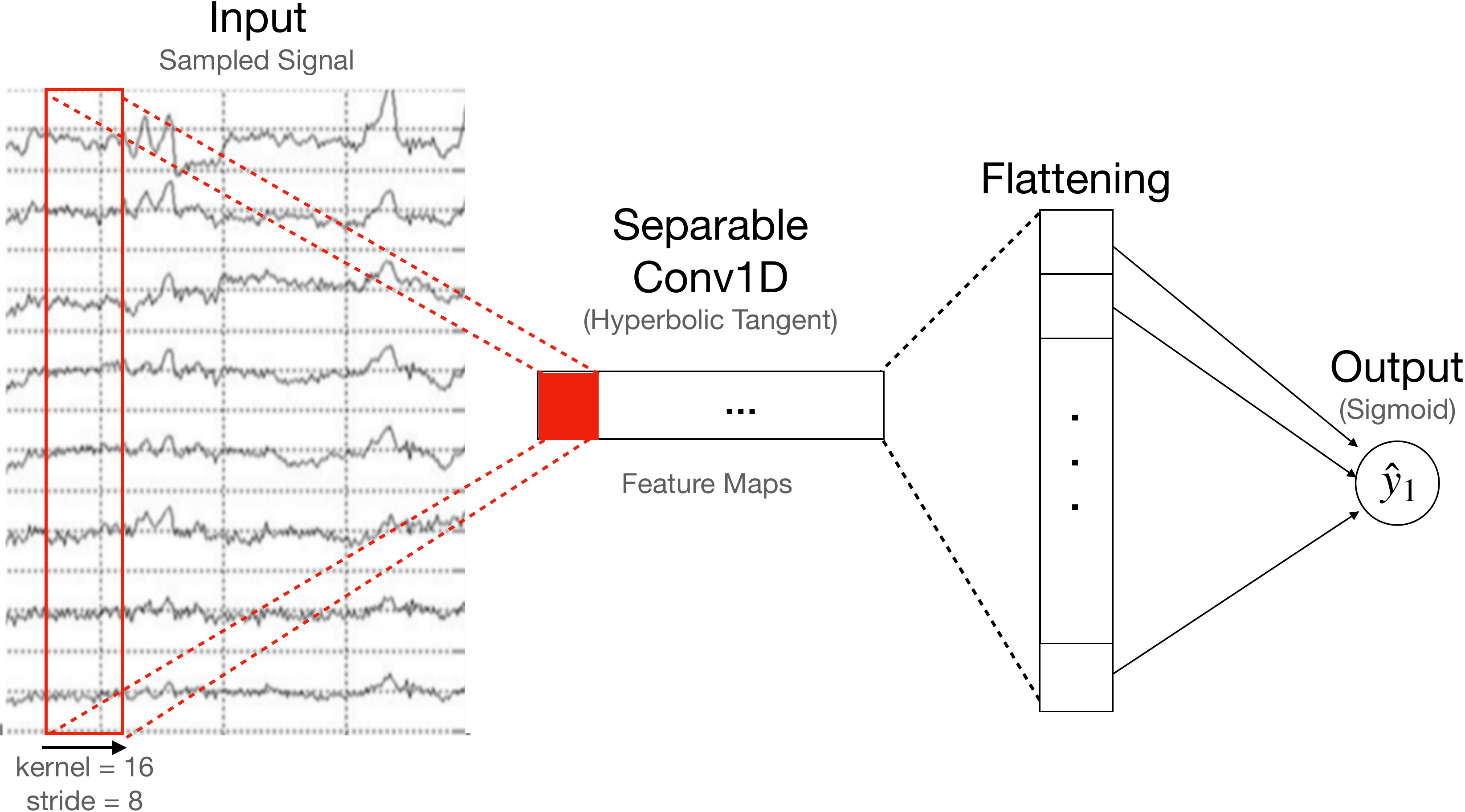}
\par\end{centering}
\caption{SepConv1D, a CNN architecture consisting of a depthwise separable 1D convolutional layer followed by a Sigmoid classification neuron. Table \ref{tab:SepConv1D-Arq-1F} shows further details about the architecture.}
\label{fig:SepConv1D}
\end{figure}

SepConv1D is a simple CNN architecture consisting of two layers. The architecture is illustrated in Figure \ref{fig:SepConv1D}. The input to SepConv1D is a matrix $\mathbf{X}^{T}$ of filtered signals. 
In the first layer, a depthwise separable 1D convolution over time with kernels of size $16 \times C$ is applied to the input. Since this operation runs on all $C$ electrodes simultaneously, this layer can learn features from both the temporal and spatial domains. We use a stride of eight for the convolution operation and a Hyperbolic Tangent Sigmoid activation function (see  \ref{sec:Appendix}).
The number of filters in the convolutional layer was set through an experimental evaluation, which is described in Section \ref{sec:Results}. Then, the classification is carried out by a single neuron with a Sigmoid activation function (see~\ref{sec:Appendix}), from the flattened output $\mathbf{h}$ of the convolutional layer. Thus, the output of SepConv1D can be expressed as:

\begin{align}
\hat{y}_1 &=\sigma(\mathbf{w}^\top\mathbf{h} + b),
\end{align}

\noindent where $\mathbf{w}$ and $b$ are the weights and bias of the output neuron respectively.

SepConv1D is inspired by OCLNN but uses a depthwise separable 1D convolutional layer instead of a standard 1D convolutional layer. The reason of using a depthwise separable convolutional layer is that it can reach a similar performance with fewer parameters and a lower computational cost compared to a standard convolutional layer~\citep{chollet2017xception, howard2017mobilenets, sandler2018mobilenetv2}.
In fact, in the P300 detection, EEGNet~\citep{Lawhern2018} has successfully exploited this type of convolutional layer to reduce the number of parameters and obtain a high detection rate.
Another difference between the two architectures is that OCLNN connects two neurons to a Softmax function in the output layer and applies Dropout, while SepConv1D connects a single neuron to a Sigmoid function and does not apply Dropout. Table \ref{tab:SepConv1D-Arq-1F} details the number of filters, the input and output sizes of the layers, and the number of parameters.  

\subsection{FCNN}  
\begin{figure}
\begin{centering}
\includegraphics[scale=0.3]{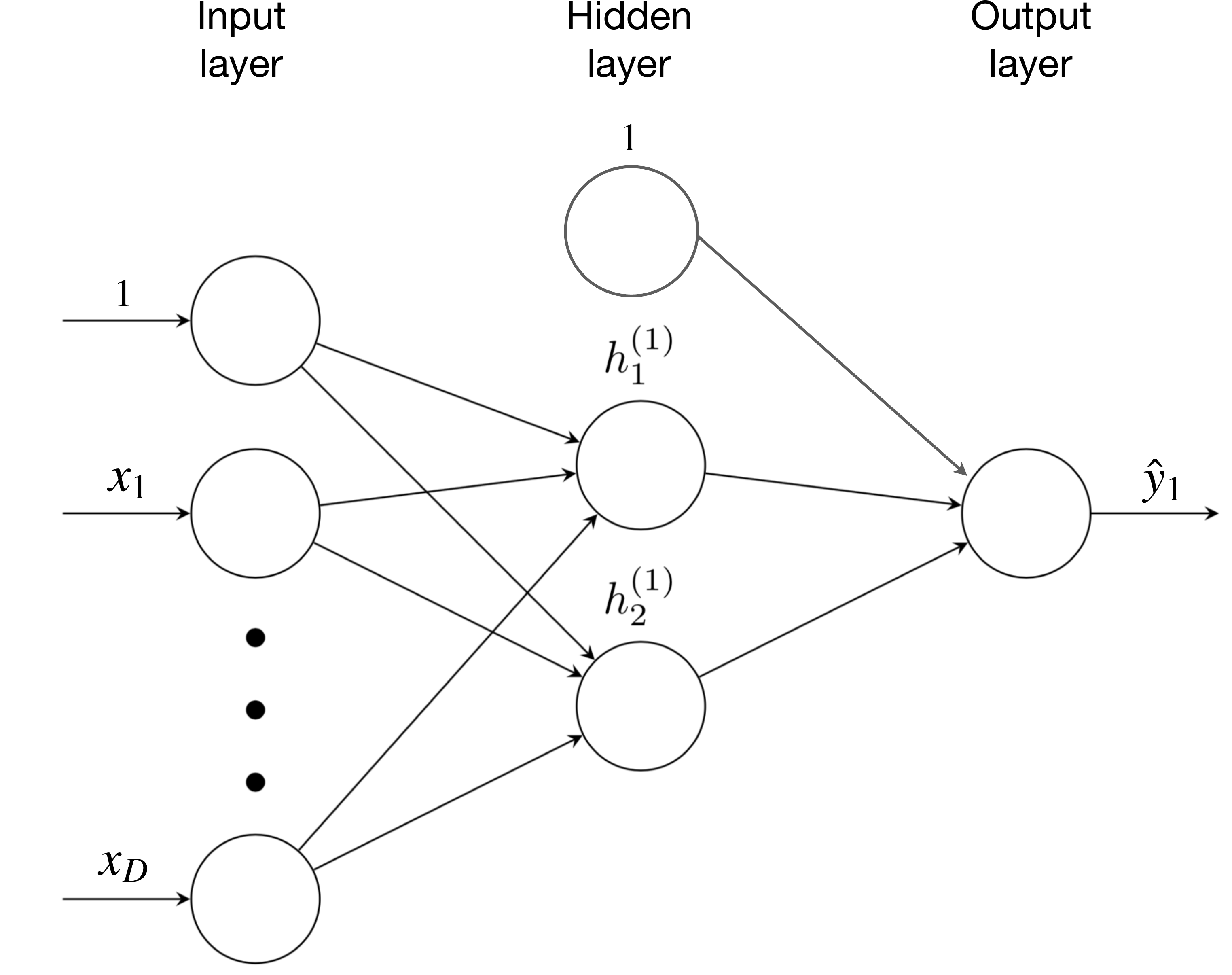}
\par\end{centering}
\caption{One-layer Fully-Connected Neural Network composed of an input, a two-neuron hidden layer, and an output layer.}
\label{fig:FCNN}
\end{figure}

We also propose the use of a one-layer Fully-Connected Neural Network (FCNN), similar to the one presented by \citep{Cechovic2013} (see Figure \ref{fig:FCNN}). We used the Hyperbolic Tangent activation function in  the hidden layer and a Sigmoid function in the output layer (see  \ref{sec:Appendix}). FCNN is composed of two neurons in the hidden layer, whose output is given by $h_j = \tanh\left(\sum_{i=1}^{D}w_{ij}^{(1)}x_{i}+b_j^{(1)}\right)$, where $w_{ij}^{(1)}$ is the connection weight from input \emph{$i$} to the intermediate neuron \emph{$j$}, $b_j^{(1)}$ is the bias of the neuron $j$, $x_i$ is an element of the vector resulting from flattening the matrix of filtered signals $\mathbf{X}$, and $D=C \cdot T$. Thus, the output of FCNN is $\hat{y}_1=\sigma \left(\sum_{j=1}^{2}w_{j1}^{(2)}h_j+b^{(2)}\right)$, where $w_{j1}^{(2)}$ is the connection weight from neuron $j$ in layer (1) to the output neuron and $b^{(2)}$ is the bias of the output neuron. See Table \ref{tab:FCNN_Arq} for more details about the architecture. 

\section {Experimental Design} \label{sec:Experimental_Design}
We compare the performance of state-of-the-art CNN architectures with SepConv1D and FCNN for both single-trial within-subject and cross-subject P300 detection. In what follows, we explain in detail the experimental design.

\subsection{Datasets} \label{sec:datasets}
For our experiments, we used four benchmark datasets where the subjects were visually stimulated with the Donchin $6 \times 6$ speller matrix described in~\citep{Donchin2000}. The speller matrix is composed of alphanumeric symbols that allow the subjects to write a word. In all datasets, the presence of the P300 component in an EEG signal was indicated by $y=1$ and the absence by $y=0$. 
\subsubsection*{P300-LINI \citep{Ledesma2010}}
This dataset is composed of the EEG signals of 22 healthy students from 21 to 25 years old without known neurological damage. The P300-LINI dataset contains signals of ten EEG channels (Fz, C4, Cz, C3, P4, Pz, P3, PO8, Oz, and PO7) and was acquired following the international 10-20 system, with the right earlobe and the right mastoid serving as reference and ground locations respectively. However, we only used six channels (Fz, Cz, Pz, PO8, Oz, and PO7), since it has been previously reported \citep{Alvarado2016b} to be the most relevant to detect the P300 component. The signal was digitized at a rate of 256 Hz and processed online using a notch filter (Chebyshev of order 4) with cutoff frequencies between 58 and 62 Hz and a bandpass filter (Chebyshev of order 8) with cutoff frequencies between 0.1 and 60 Hz.

In order to generate the oddball paradigm, the rows and columns of the matrix flashed randomly 15 times every 125 ms (i.e., trials); each flash lasted 62.5 ms. The resulting set ${D}_1$ consists of 480 EEG signals that potentially contained P300 and 2,400 EEG signals without P300, for each subject. 
Each segment was filtered offline using a 4th-order Butterworth bandpass filter with bandwidth range from 0.1 to 12 Hz to extract the ERP signals embedded in the EEG. The DC component was removed by subtracting the mean of each electrode from the filtered signal. We extracted segments of 800 ms of EEG data after every stimulus. Finally, the linear trend was removed from each segment.

\subsubsection*{BCI Competitions}
BCI Competition II - Data set IIb \citep{Blankertz2004}  and BCI Competition III - Data set II \citep{Blankertz2006} contain 64 channels and were acquired following the international 10-10 system. The signals were digitized at a rate of 240 Hz and filtered with cutoff frequencies between 0.1 and 60 Hz.  In order to generate the oddball paradigm, the rows and columns of the matrix flashed randomly 15 times every 100 ms, with an inter stimulus interval of 75 ms. 
 
BCI Competition II - Data set IIb \citep{Blankertz2004} is composed of the EEG signals of one subject collected in three sessions (sessions 10-12): session 10 has five runs, session 11 has six runs, and session 12 has eight runs. We only processed  the labeled data of sessions 10 and 11. In each case, we extracted segments of 650 ms of EEG data after every stimulus. The resulting set ${D}_2$ consists of 570 EEG signals that potentially contained P300 and 2,850 EEG signals without P300. 

BCI Competition III - Data set II \citep{Blankertz2006} is composed of the EEG signals of two subjects. Each subject generated a training and a test set. We processed the training sets of both subjects since they are labeled. We extracted segments of 1000 ms of EEG data after every stimulus. The resulting set ${D}_3$ consists of 3,825 EEG signals that potentially contained P300 and 12,750 EEG signals without P300, for each subject.

\subsubsection*{BNCI Horizon 2020}

Dataset 8. P300 speller with ALS patients was acquired by Riccio et al.  \citep{Riccio2013} and its available at \citep{BNCI_Horizon2014}. This dataset is composed of the EEG signals of eight subjects with Amyotrophic Lateral Sclerosis. This dataset contains eight EEG channels (Fz, Cz, Pz, Oz, P4, P3, PO8,  and PO7) and was acquired following the international 10-10 system. All channels were referenced to the right earlobe and grounded to the left mastoid. The signal was digitized at a rate of 256 Hz and band-pass filtered with cutoff frequencies between 0.1 and 30 Hz. In order to generate the oddball paradigm, the rows and columns of the matrix flashed randomly 10 times at a rate of 4 Hz for 125 ms, with an inter stimulus interval of 125 ms. We extracted segments of 800 ms of EEG data after every stimulus. The resulting set ${D}_4$ consists of 700 EEG signals that potentially contained P300 and 3,500 EEG signals without P300, for each subject.

\subsection{Evaluation}
For the architectures under analysis, we ran 200 training iterations (epochs) of the Adam optimizer with the parameters recommended by \citep{Kingma2015} and performed early stopping if the validation loss did not improve in 50 iterations. Binary Cross-Entropy was used as loss function for BN$^3$, FCNN and SepConv1D, Mean Squared Error for CNN1 and CNN3, and Categorical Cross-Entropy for the rest of the architectures.
The Glorot uniform initializer was employed for all architectures, except for CNN1, UCNN1, CNN3 and UCNN3, which employed the initializer proposed by Cecotti and Gräser~\citep{Cecotti2011}. In addition, we standardize each EEG channel separately. 

 For within-subject detection, we performed 10 repetitions of stratified 5-fold cross-validation for each subject. In each repetition 5 splits were generated using 5-fold cross-validation, where a split consisted of a training set with 80\% of the data of a given subject and a validation set with the remaining 20\%.  For cross-subject detection, we used data from a subset of subjects to train a model for another subject. Consequently, the evaluation of this type of detection was only possible on the $D_1$ and $D_4$ datasets, given that the $D_2$ and $D_3$ datasets have only 1 and 2 subjects respectively. For this evaluation, we performed a leave-two-out cross-validation, where one subject is selected for testing, another for validation, and the remaining for training. This process was repeated for each subject, producing 22 folds for $D_1$ and 8 folds for $D_4$. For both types of detection, we computed the Area Under the Curve (AUC) of the Receiver Operating Characteristic (ROC) over the test set in each split.

\subsection{Implementation details}
The experiments were performed on a single PC with Linux Ubuntu 16.04, an Intel(R) Core(TM) i7-6700K CPU @ 4.00GHz, 64 GB in RAM, and an NVIDIA GeForce GTX 1080 GPU with 2560 CUDA cores and 8 GB of RAM. The architectures were implemented in Keras~\citep{Chollet2015} with Tensorflow 1.14.0~\citep{Abadi2015} as backend. For DeepConvNet, ShallowConvNet and EEGNet, we used the code provided by Lawhern et al.~\citep{Lawhern2018}:~\url{https://github.com/vlawhern/arl-eegmodels}. The rest of the architectures were implemented following the descriptions of the corresponding papers. The source code for the architectures under analysis and the reported experiments is available at~\url{http://github.com/gibranfp/P300-CNNT}.

\begin{figure*}[t!]
    \centering
    \begin{subfigure}[t]{0.5\textwidth}
        \centering
        \includegraphics[height=1.2in]{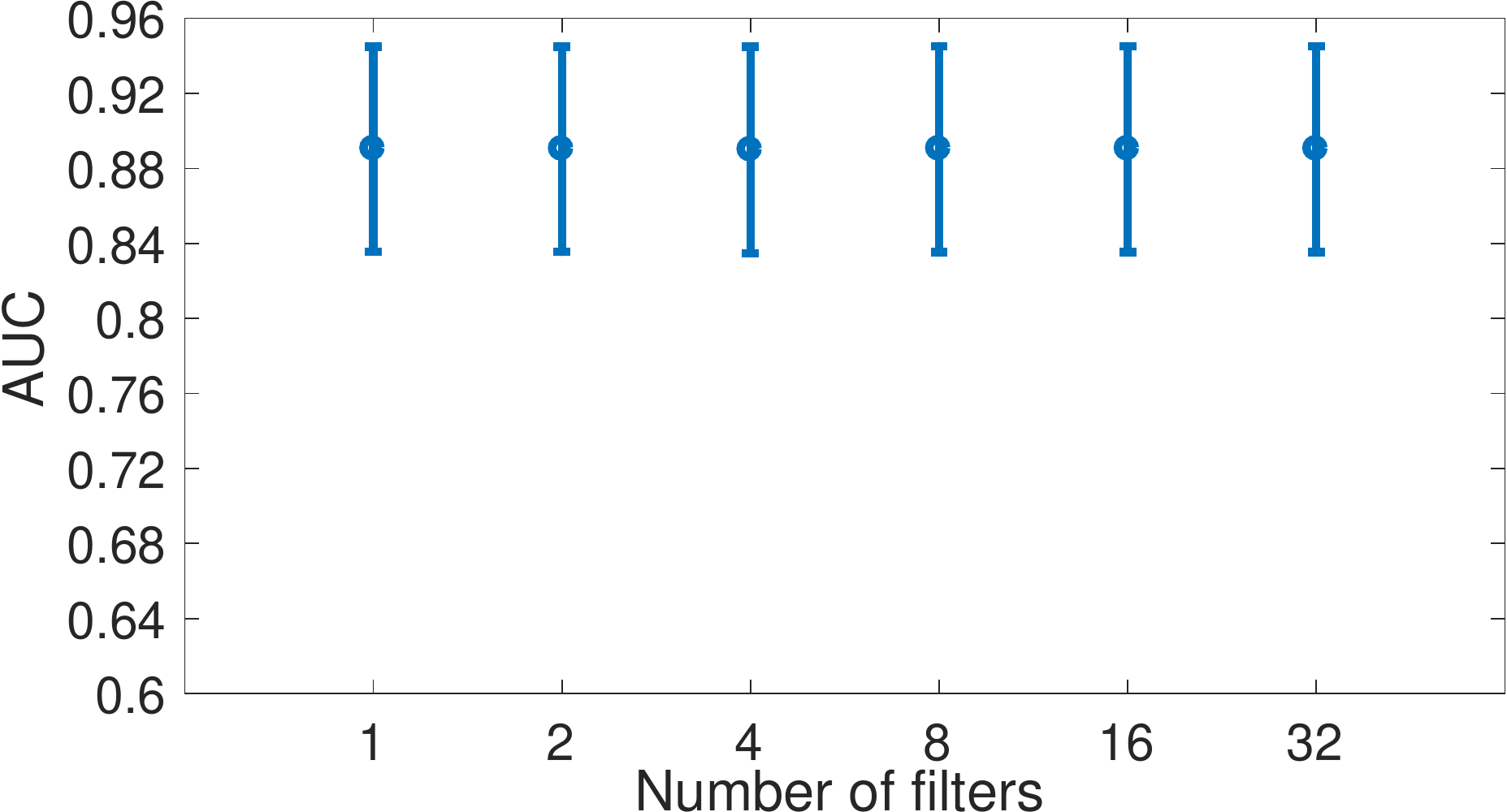}
        \caption{${D}_1$}
    \end{subfigure}%
    ~ 
    \begin{subfigure}[t]{0.5\textwidth}
        \centering
        \includegraphics[height=1.2in]{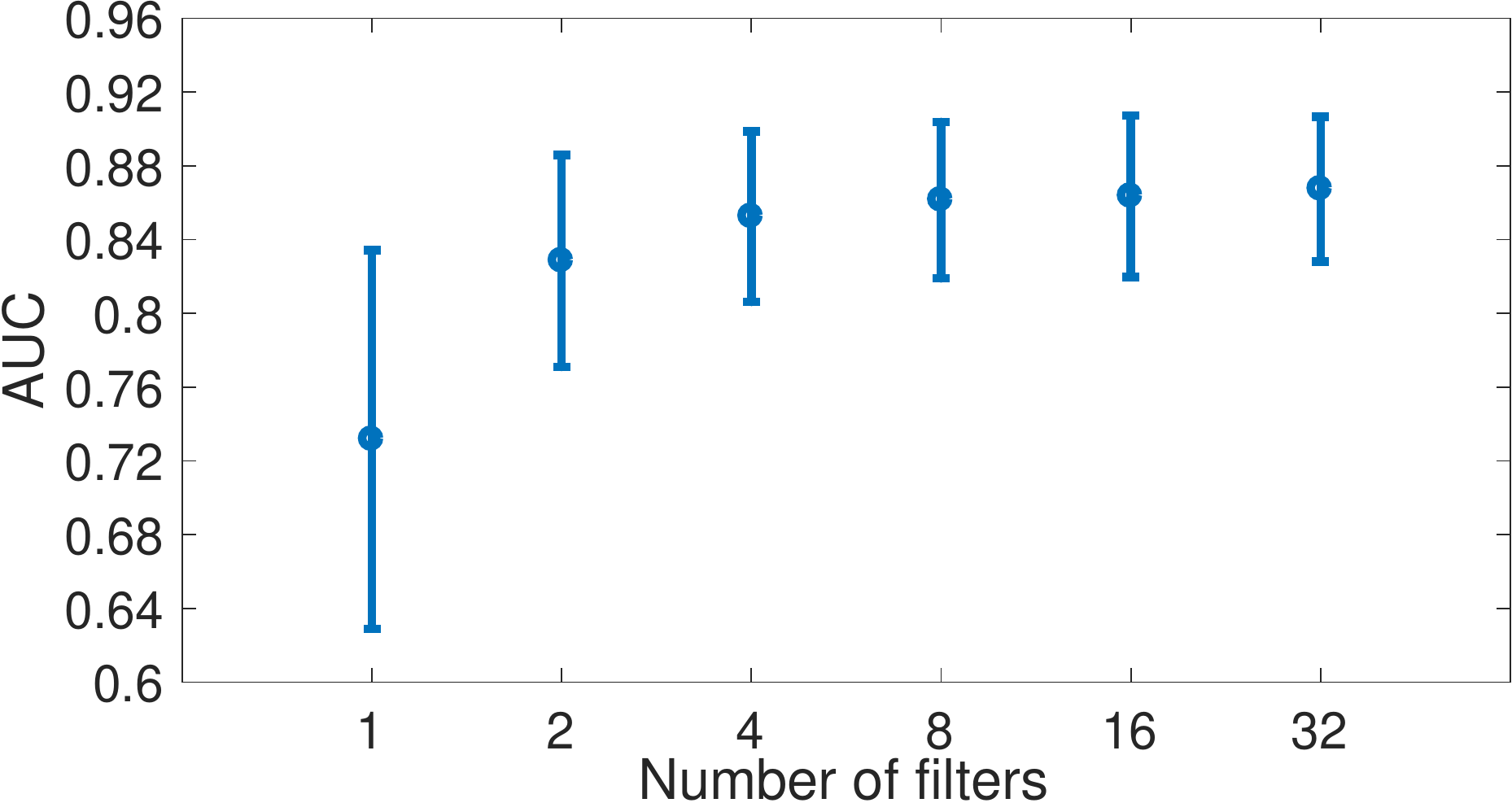}
        \caption{${D}_2$}
    \end{subfigure}
           
     \begin{subfigure}[t]{0.5\textwidth}
        \centering
        \includegraphics[height=1.2in]{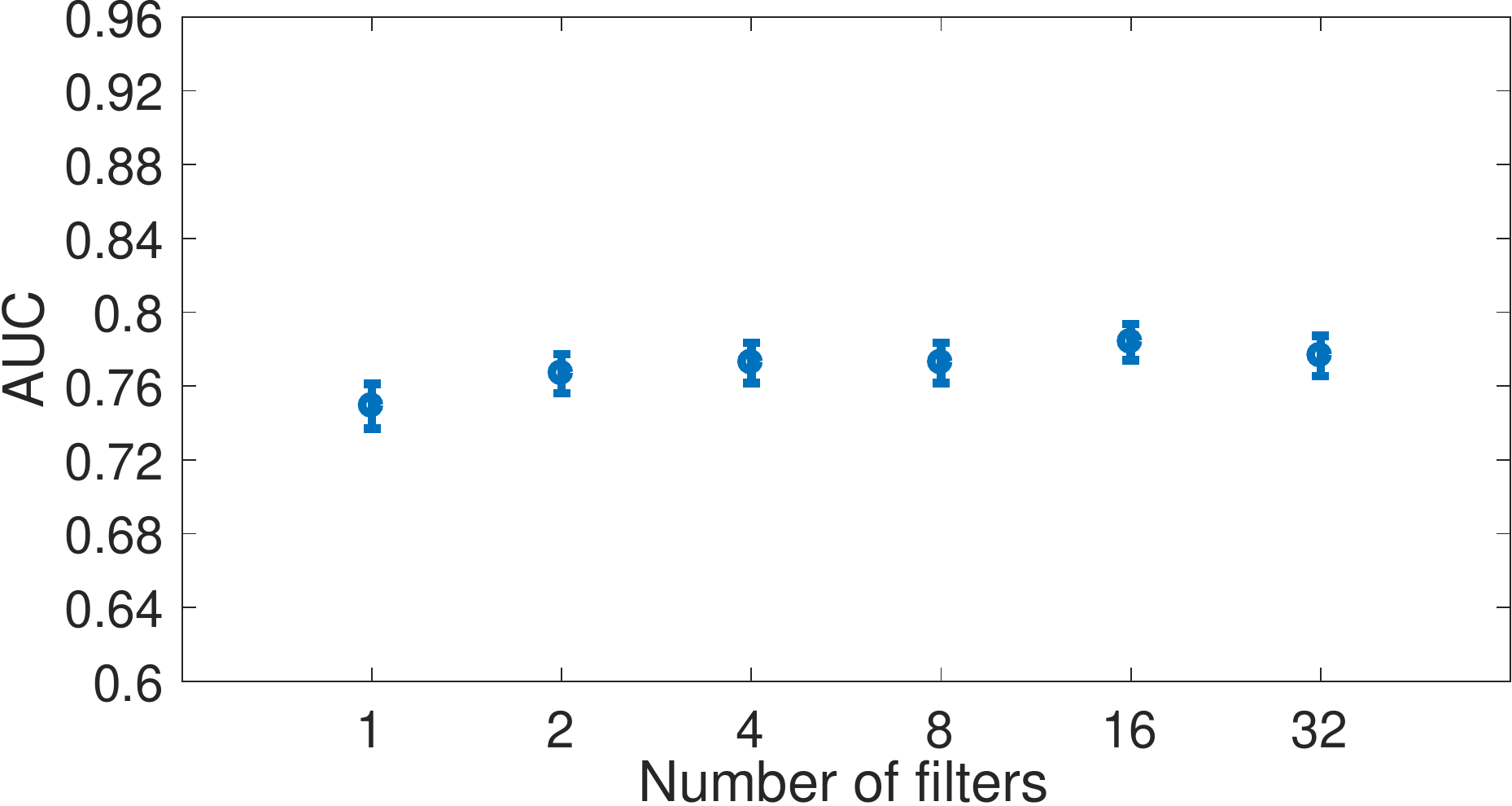}
        \caption{${D}_3$}
    \end{subfigure}%
    ~ 
    \begin{subfigure}[t]{0.5\textwidth}
        \centering
        \includegraphics[height=1.2in]{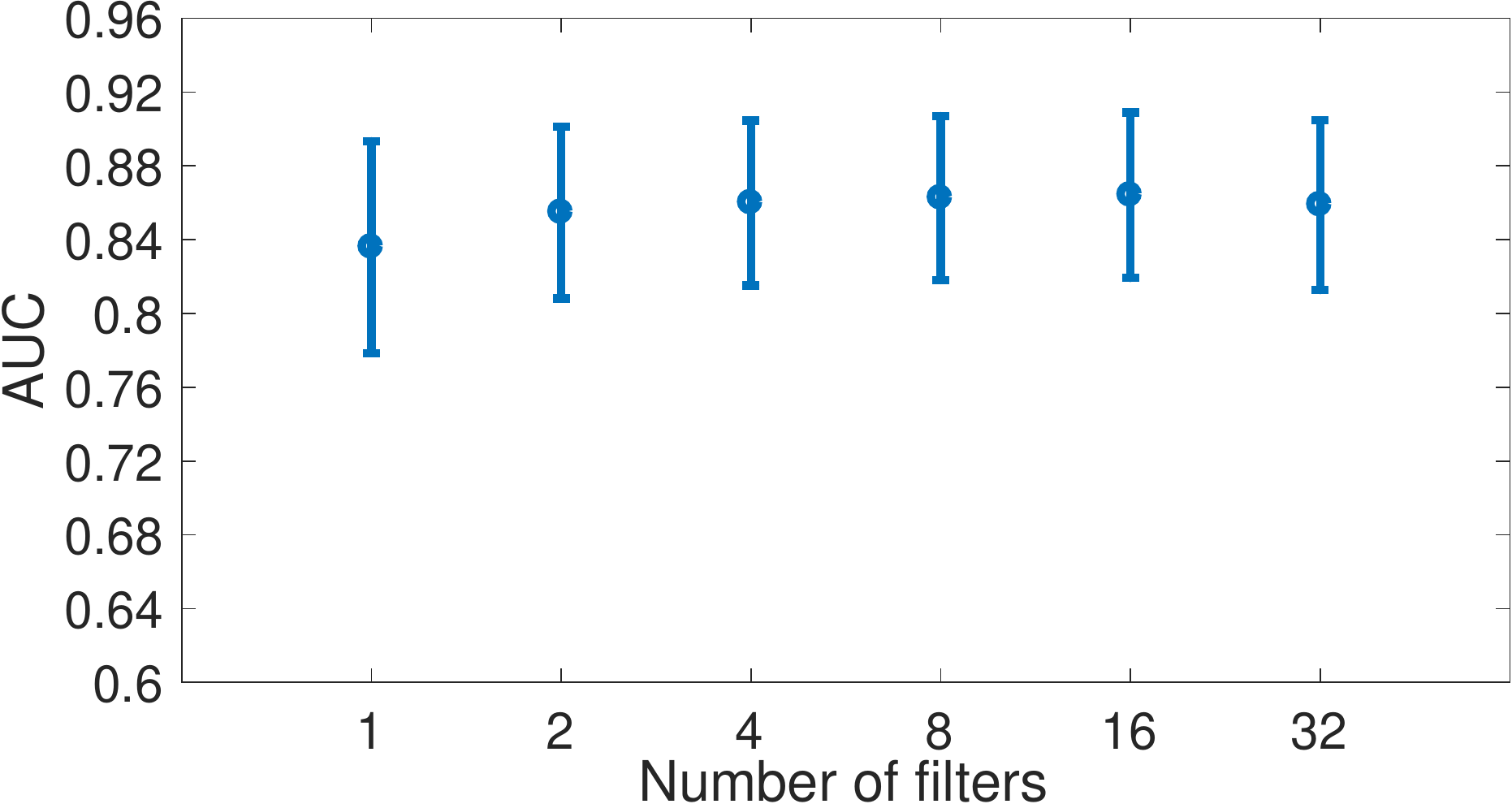}
        \caption{${D}_4$}
    \end{subfigure}    
\caption{Comparison of SepConv1D mean AUC with different number of filters for within-subject P300 detection.}
\label{fig:NumbFilters_within}
\end{figure*}

\section {Results and Discussion}\label{sec:Results}

\subsection{Within-subject classification} 
 In this section, we describe the experiments for within-subject P300 detection. These experiments were performed on the four benchmark datasets described in Section \ref{sec:datasets}.

\subsubsection{Selection of the number of filters for the SepConv1D} \label{sec:Filters_With}
In order to find a trade-off between computational complexity and detection performance, we evaluated SepConv1D with 1, 2, 4, 8, 16 and 32 filters. Figure~\ref{fig:NumbFilters_within} shows the mean and standard deviation of the AUC values obtained over all repetitions, splits and subjects for SepConv1D with different number of filters on the four benchmark datasets. For ${D}_1$ the performance is almost identical in all cases, while for ${D}_2$ the performance is significantly lower with 1 filter, slightly lower with 2 filters and very similar from 4 to 32 filters. For ${D}_3$ the highest mean AUC is achieved with 16 filters, although the difference is very small compared with less or more filters.
Finally, for $D_4$ the mean AUC values are slightly lower with 1 filter and very similar with larger number of filters.
Given these results, 4 filters seems to offer a good trade-off for all datasets. Thus, we decided to use SepConv1D with four filters for subsequent experiments; the amount of trainable parameters for this configuration is as follows (see Table \ref{tab:Parametros}): $225$ for $D_1$, $1,361$ for ${D}_2$, $1,405$ for ${D}_3$, and $265$ for ${D}_4$. See Table \ref{tab:SepConv1D-Arq-1F} for more details about the SepConv1D architecture with four filters and EEG signals of six channels and 206 samples as input.

\begin{table}
\centering
\begin{threeparttable}
\scriptsize
\begin{tabular}{|l|p{0.7cm}|p{1.7cm}|p{1.2cm}|p{1cm}|p{1.4cm}|p{1.2cm}|}
\hline
\textbf{Layer} &  \textbf{No. filters} & \textbf{Size} & \textbf{No. params} & \textbf{Output}  &\textbf{Activation function} & \textbf{Options} \\
\hline\hline
Input & & $ 206 \times 6$ & & & &\\
ZeroPadding1D &  &  &  & $(214,6)$ & & padding = 4 \\ 
SeparableConv1D & 4 & kernel = 16, stride = 8 & 124 & $(25,4)$ & & \\ 
Activation & & & & $(25,4)$ & $tanh$ &  \\
Flatten	& & & &$(25)$ & &\\
Dense & 1 & & 101 & $(1)$ & Sigmoid & \\\hline
  \end{tabular}
\end{threeparttable}
\caption{SepConv1D architecture for within-subject classification with four filters  for dataset ${D}_1$ (see Section~\ref{sec:datasets}).}
\label{tab:SepConv1D-Arq-1F}
\end{table}

\begin{table}[bth!]
\centering
\begin{tabular}{|c |c |c | c | c |} 
 \hline
 Architecture & ${D}_1$ & ${D}_2$& ${D}_3$ & ${D}_4$ \\
 \hline\hline
CNN1  &0.89$\pm$0.06&0.88$\pm$0.04&0.82$\pm$0.04&0.85$\pm$0.05\\
UCNN1 &0.88$\pm$0.07&0.87$\pm$0.08&0.81$\pm$0.05&0.85$\pm$0.05\\
CNN3  &0.72$\pm$0.16&0.78$\pm$0.14&0.68$\pm$0.13&0.66$\pm$0.14\\
UCNN3 &0.77$\pm$0.13&0.81$\pm$0.09& 0.7$\pm$0.11&0.71$\pm$0.13\\
CNNR  &0.88$\pm$0.06&0.86$\pm$0.05&0.76$\pm$0.06&0.86$\pm$0.04\\
DeepConvNet    & 0.9$\pm$0.05& 0.9$\pm$0.04&0.84$\pm$0.04&0.88$\pm$0.04\\
ShallowConvNet &0.83$\pm$0.08&0.82$\pm$0.07&0.78$\pm$0.05&0.86$\pm$0.05\\
BN$^3$&0.88$\pm$0.06&0.81$\pm$0.05& 0.8$\pm$0.05&0.84$\pm$0.05\\
EEGNet&0.89$\pm$0.05& 0.9$\pm$0.04&0.83$\pm$0.04&0.88$\pm$0.04\\
OCLNN&0.89$\pm$0.05&0.85$\pm$0.05& 0.8$\pm$0.05&0.85$\pm$0.05\\
FCNN &0.89$\pm$0.05&0.78$\pm$0.05&0.75$\pm$0.04&0.84$\pm$0.05\\
SepConv1D &0.88$\pm$0.06&0.85$\pm$0.05&0.82$\pm$0.04&0.86$\pm$0.05\\
 \hline
\end{tabular}
\caption{Mean AUC and standard deviation obtained by the architectures under analysis for within-subject single-trial P300 detection on all datasets. See details in tables \ref{tab:AUCdetail-within-lini}-\ref{tab:AUCdetail-within-als}. All the results are rounded up to two decimal places.}
\label{tab:AUC_within}
\end{table}

\subsubsection{Comparison with state-of-the-art CNN-based architectures}\label{ComparisonWithin}
Table \ref{tab:AUC_within} compares the performance of SepConv1D and FCNN with state-of-the-art CNN architectures for within-subject P300 detection on the four benchmark datasets. The mean and standard deviation of the AUC values obtained over all repetitions, splits and subjects are reported for each architecture and dataset. In general, we can observe higher mean AUC values (but also higher standard deviations) in $D_1$, and the difference in performance between architectures is smaller. In contrast, lower AUC values are observed in $D_3$.  

Among the evaluated architectures, DeepConvNet achieved the top performance in all datasets, followed by EEGNet, whereas the lowest performances were obtained by CNN3 and UCNN3.

 \begin{figure*}[t!]
    \centering
    \begin{subfigure}[t]{0.5\textwidth}
        \centering
        \includegraphics[height=1.75in]{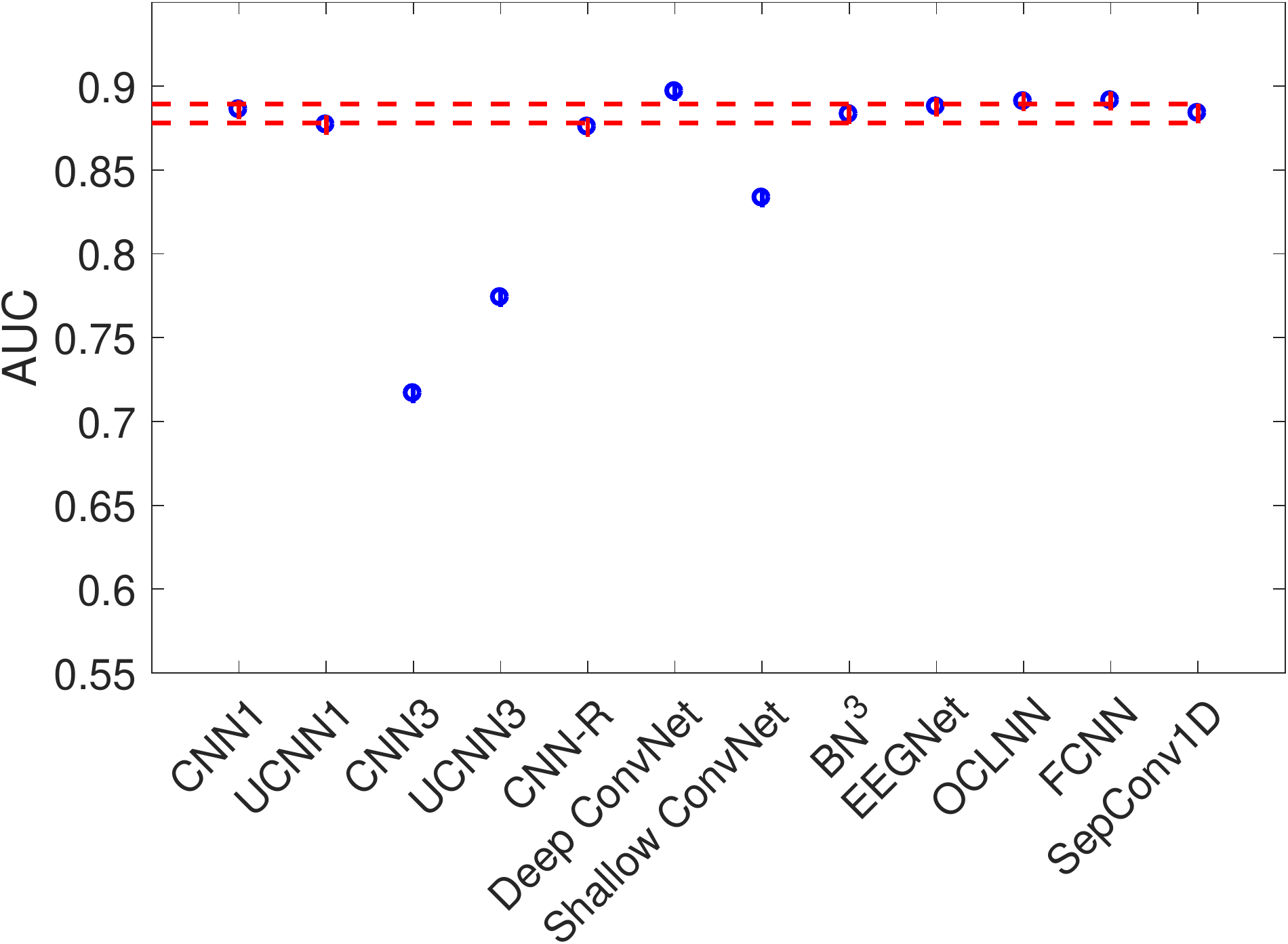}
        \caption{${D}_1$}
    \end{subfigure}%
    ~ 
    \begin{subfigure}[t]{0.5\textwidth}
        \centering
        \includegraphics[height=1.75in]{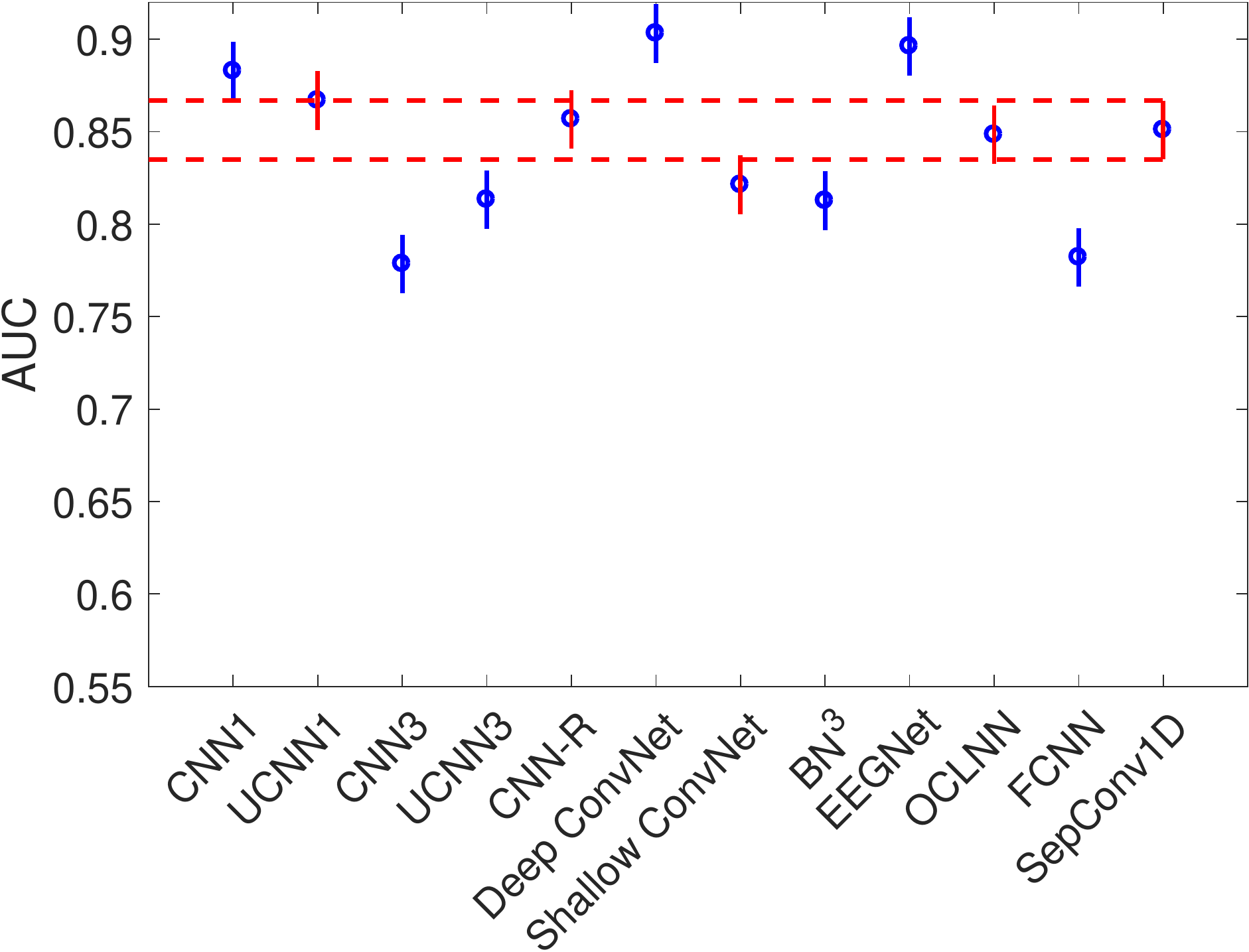}
        \caption{${D}_2$}
    \end{subfigure}
           
     \begin{subfigure}[t]{0.5\textwidth}
        \centering
        \includegraphics[height=1.75in]{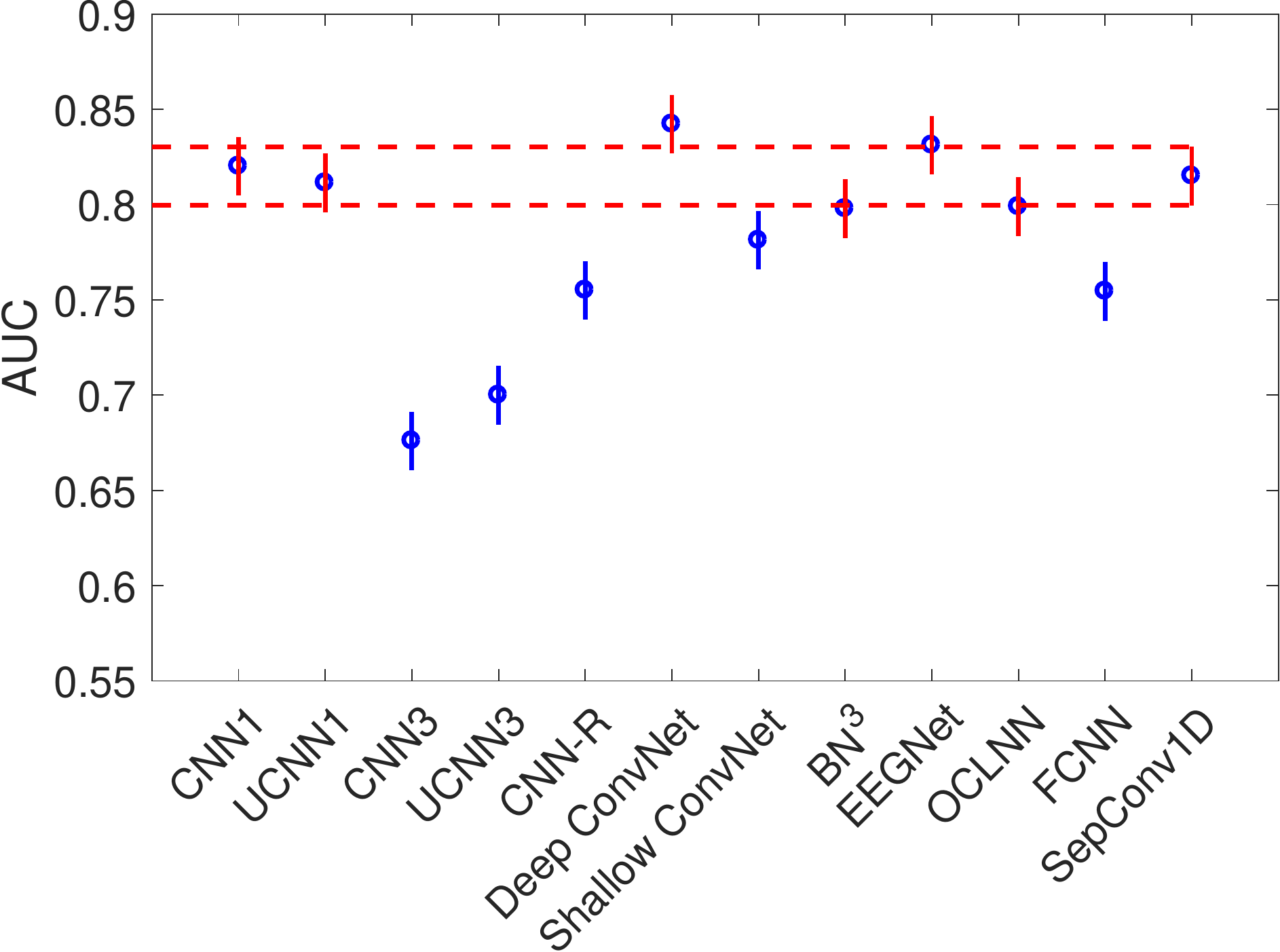}
        \caption{${D}_3$}
    \end{subfigure}%
    ~ 
    \begin{subfigure}[t]{0.5\textwidth}
        \centering
        \includegraphics[height=1.75in]{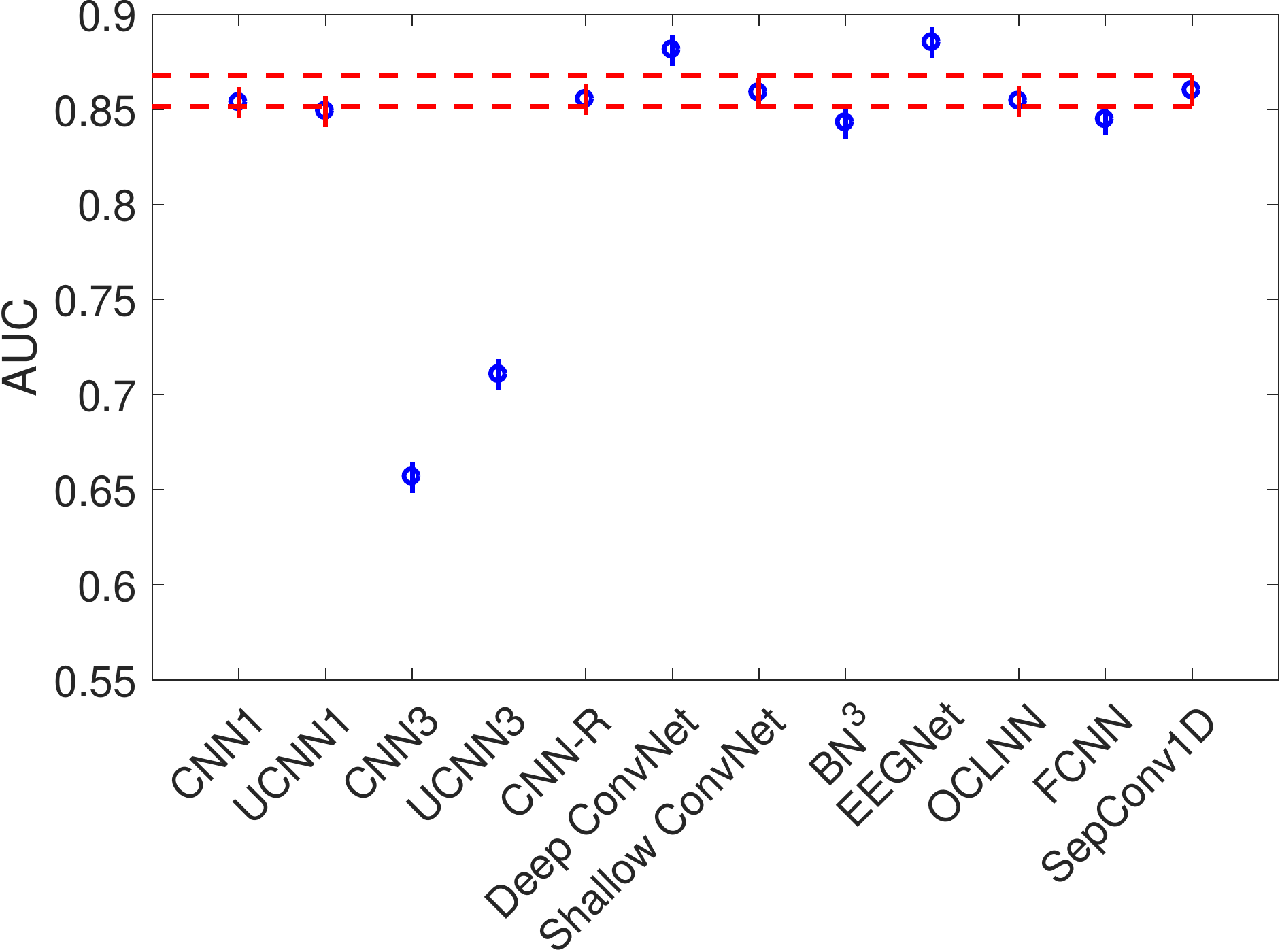}
        \caption{${D}_4$}
    \end{subfigure}    
\caption{Comparison between the AUC values obtained by SepConv1D against those of state-of-the-art CNN architectures for within-subject P300 detection on the four benchmark datasets. The red dotted lines indicate the minimum and maximum values of the AUC's standard deviation corresponding to SepConv1D.}
\label{fig:Multicomparison}
\end{figure*}
 
For each dataset, we tested the null hypothesis that the AUC values were equal between Sepconv1D and the other architectures under analysis. To that end, we performed a statistical testing using a one-way analysis of variance (ANOVA). The results are shown  in Figure \ref{fig:Multicomparison}. The red dotted lines represent the intervals for the mean AUC values for SepConv1D. The AUC values that lie inside the red dotted lines are those where the differences with the AUC values of SepConv1D are not statistically significant. As can be seen, the AUC values obtained by SepConv1D are close to those obtained by the top performing architectures on all datasets. Specifically, the difference with DeepConvNet is statistically significant for $D_1$, $D_2$ and $D_4$, and with EEGNet only for $D_2$ and $D_4$. There is no statistically significant difference with the top performing architectures for $D_3$. In all cases, the differences between UCNN1, OCLNN, and SepConv1D are not statistically significant. On the other hand, the performance of FCNN was significantly lower than most architectures for $D_2$ and $D_4$ but was relatively competitive for $D_1$ and $D_4$. CNN1 also achieved higher AUC values than SepConv1D for $D_1$ and $D_2$ but the difference is statistically significant only for the latter.  
\subsection{Cross-subject P300 detection}
 We now proceed to describe the experiments for cross-subject P300 detection. These experiments were performed only on datasets ${D}_1$ and ${D}_4$ because $D_2$ and $D_3$ had only one and two subjects respectively.

\subsubsection{Selection of the number of filters for the SepConv1D}
\begin{figure*}
    \centering
    \begin{subfigure}[t]{0.5\textwidth}
        \centering
        \includegraphics[height=1.2in]{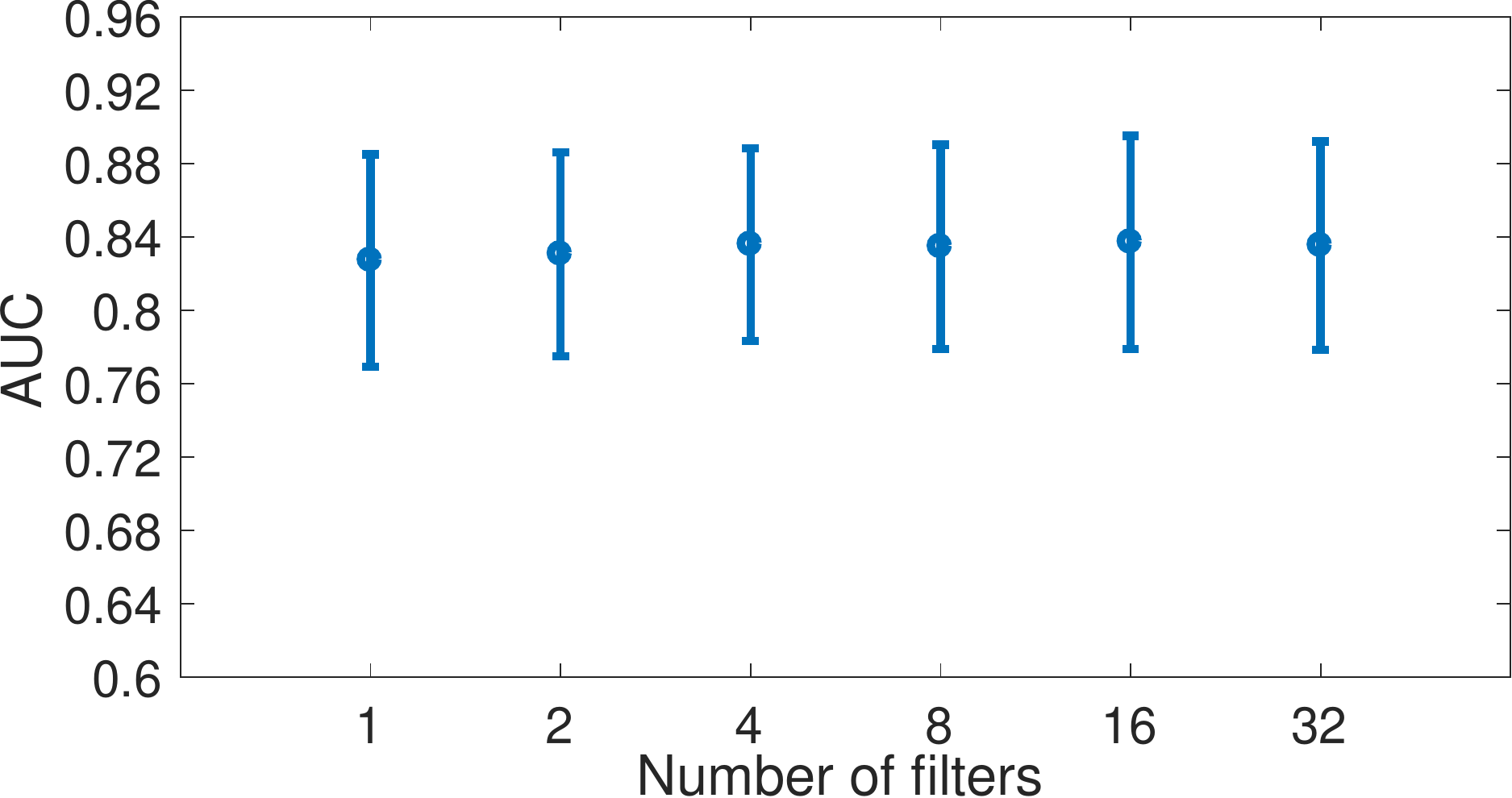}
        \caption{${D}_1$}
    \end{subfigure}%
    ~ 
    \begin{subfigure}[t]{0.5\textwidth}
        \centering
        \includegraphics[height=1.2in]{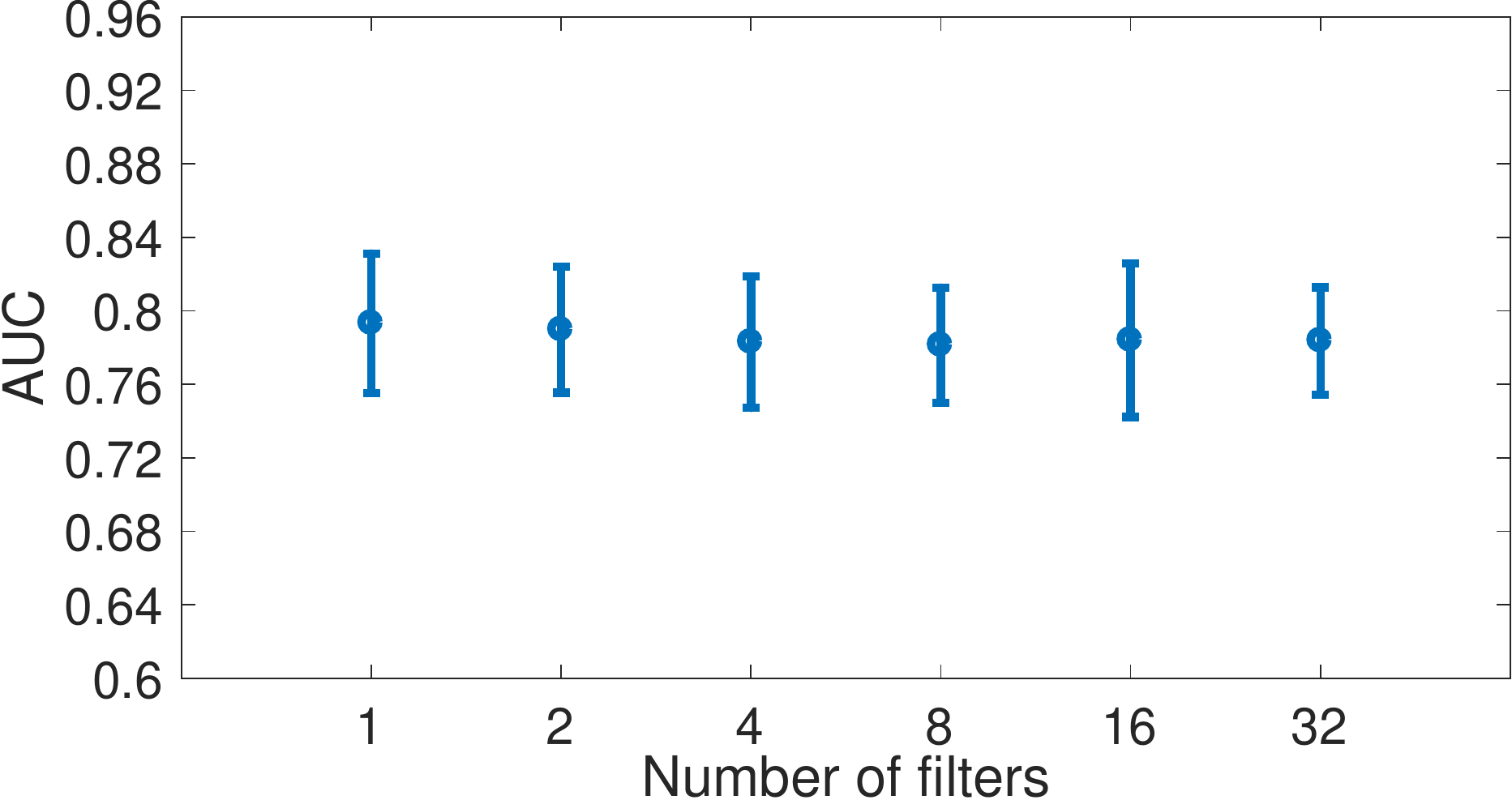}
        \caption{${D}_4$}
    \end{subfigure}
   
\caption{Comparison of mean AUC values obtained by SepConv1D with different number of filters for cross-subject P300 detection on datasets (a) ${D}_1$ and (b) ${D}_4$. }
\label{fig:NumbFilters_cross}
\end{figure*}

As in within-subject P300 detection, we carried out experiments with 1, 2, 4, 8, 16 and 32 filters for SepConv1D to find a trade-off between computational complexity and performance for cross-subject P300 detection. 
Figure \ref{fig:NumbFilters_cross} shows the mean AUC and the standard deviation for $D_1$ and $D_4$. Interestingly, the mean AUC and the standard deviation of all configurations are very similar in both datasets. For consistency, we used four filters in the subsequent experiments. 

\begin{table}[bht!]
\centering
\begin{tabular}{|c |c |c |} 
 \hline
 Architecture & ${D}_1$ & ${D}_4$ \\
 \hline\hline
CNN1 &0.82$\pm$0.05&0.78$\pm$0.04\\
UCNN1&0.84$\pm$0.06&0.78$\pm$0.05\\
CNN3&0.78$\pm$0.11&0.73$\pm$0.08\\
UCNN3&0.83$\pm$0.06&0.76$\pm$0.07\\
CNN-R&0.83$\pm$0.06&0.79$\pm$0.04\\
DeepConvNet&0.84$\pm$0.06&0.79$\pm$0.04\\
ShallowConvNet&0.82$\pm$0.07&0.79$\pm$0.03\\
BN$^3$&0.83$\pm$0.06&0.78$\pm$0.04\\
EEGNet&0.84$\pm$0.06& 0.8$\pm$0.03\\
OCLNN&0.83$\pm$0.06&0.79$\pm$0.04\\
FCNN&0.83$\pm$0.06&0.75$\pm$0.04\\
SepConv1D&0.84$\pm$0.06&0.78$\pm$0.04\\
 \hline
\end{tabular}
\caption{Mean AUC and standard deviation obtained by the architectures under analysis for single-trial cross-subject P300 detection on ${D}_1$ and ${D}_4$. See details in tables~\ref{tab:AUCdetail-cross-lini} and~\ref{tab:AUCdetail-within-als}. All the results are rounded up to two decimal places.}
\label{tab:AUC_cross}
\end{table}

\subsubsection{Comparison with state-of-the-art CNN-based architectures}
Table \ref{tab:AUC_cross} presents the performance of all architectures for cross-subject P300 detection on datasets ${D}_1$ and ${D}_4$. The mean and standard deviation of the AUC values obtained over all splits are reported for each architecture (see details in tables \ref{tab:AUCdetail-cross-lini} and \ref{tab:AUCdetail-within-als}). As expected, the cross-subject AUC values are lower than those of within-subject. However, the difference is not particularly large, so this type of detection may provide a reasonable performance compared to within-subject detection when subject calibration is not possible. Also, it may represent a great advance in the research for the generalized usage of BCI by eliminating the subject calibration. Again, we can observe higher AUC values for $D_1$ compared with those in $D_4$. In these experiments, EEGNet obtained the highest AUC values, although in general its performance is very similar to most of the other architectures. 

\begin{figure*}[tb!]
    \centering
    \begin{subfigure}[t]{0.5\textwidth}
        \centering
    \includegraphics[height=1.8in]{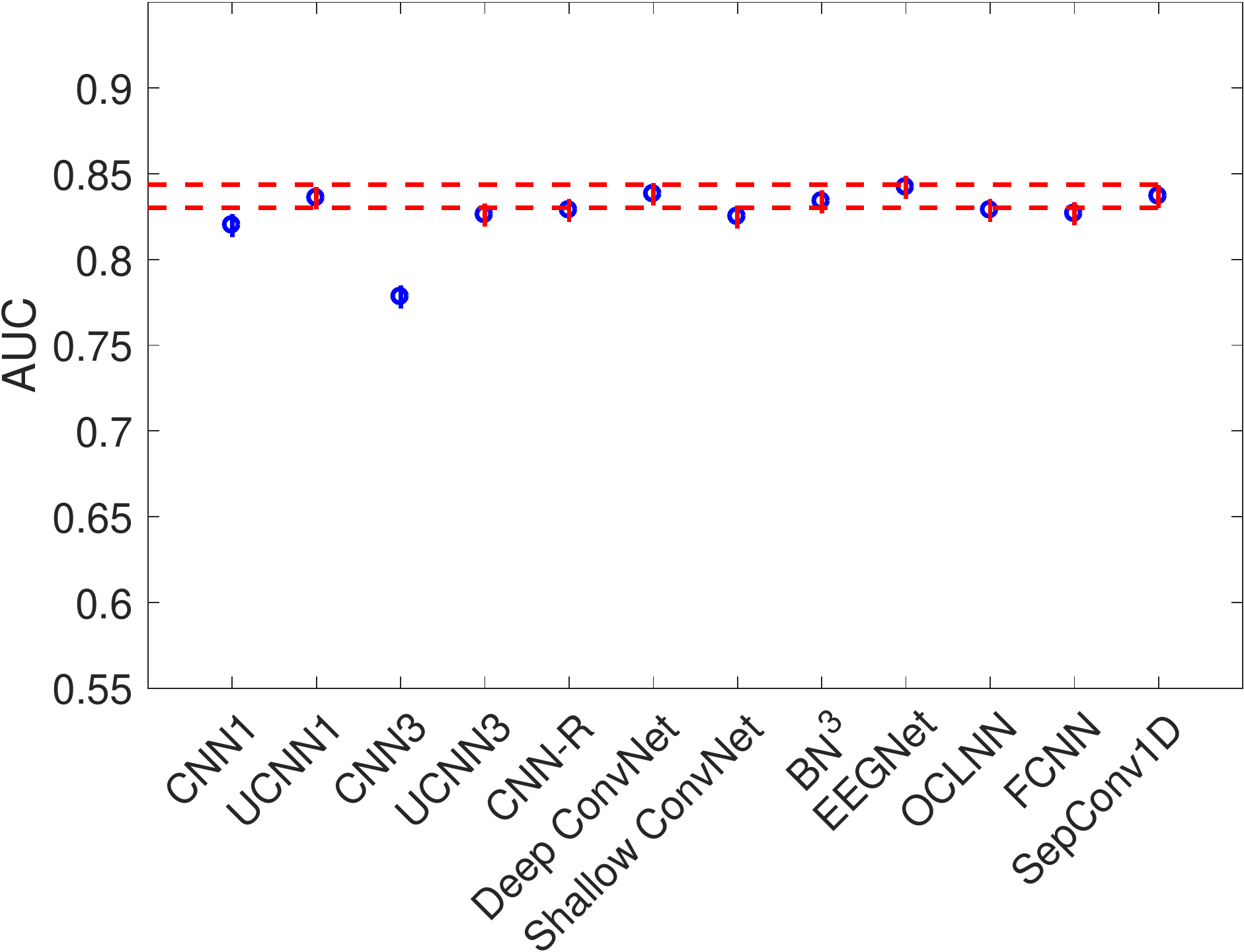}
        \caption{${D}_1$}
    \end{subfigure}%
    ~ 
    \begin{subfigure}[t]{0.5\textwidth}
        \centering
        \includegraphics[height=1.8in]{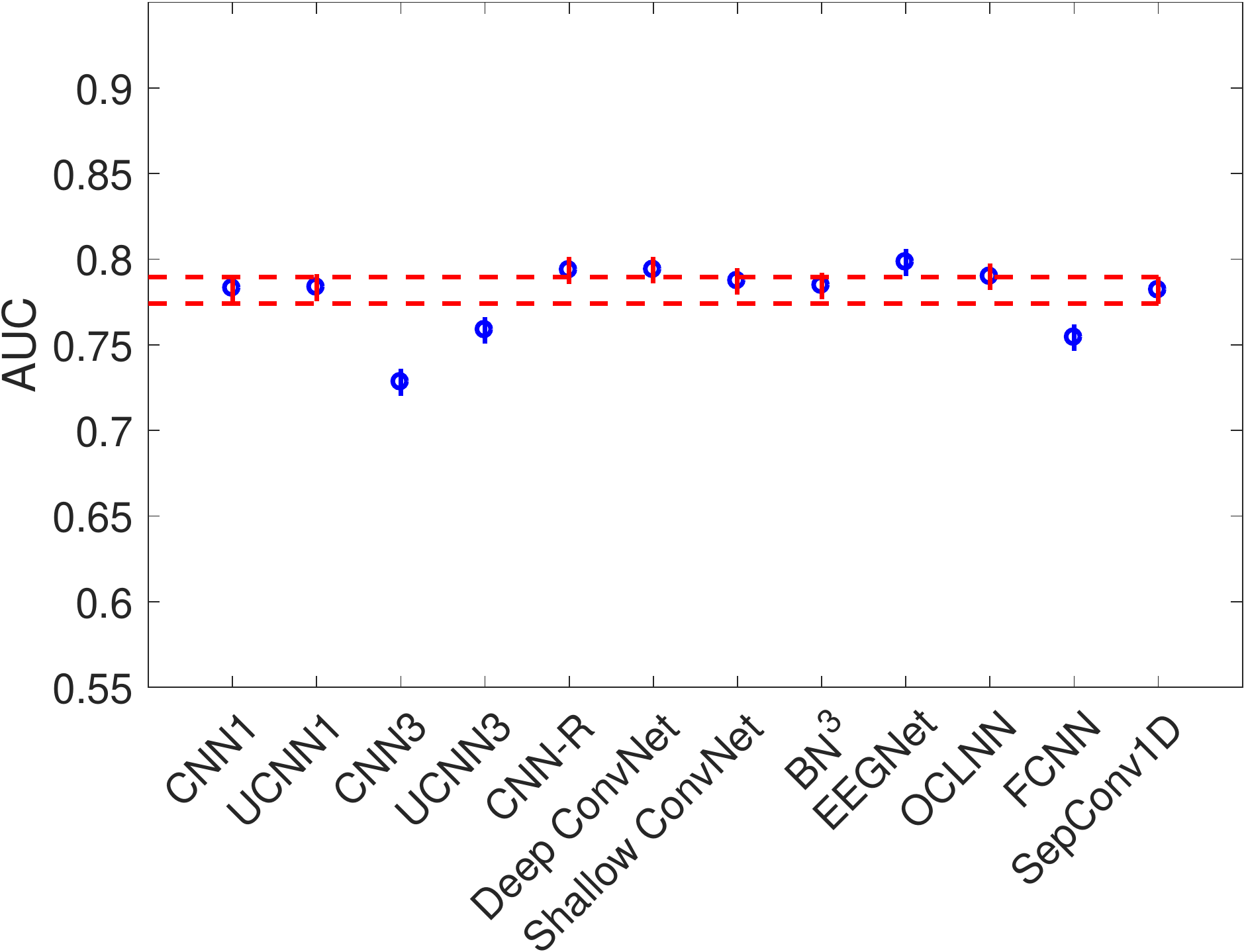}
        \caption{${D}_4$}
    \end{subfigure}
\caption{Comparison between the AUC values obtained by SepConv1D against the state-of-the-art CNN architectures for cross-subject P300 detection on ${D}_1$  and ${D}_4$. The red dotted lines indicate the minimum and maximum values of the AUC's standard deviation corresponding to SepConv1D.}
\label{fig:Multicomparison_cross}
\end{figure*}

Similarly to Section \ref{ComparisonWithin}, we performed  statistical testing using a one-way analysis of variance (ANOVA). The results are shown  in Figure \ref{fig:Multicomparison_cross}. Since almost all the architectures under  analysis lie inside the red dotted lines, we can conclude that the differences with the AUC values of SepConv1D are not statistically significant; except for CNN1 and CNN3 on ${D}_1$, and CNN3, UCNN3, EEGNet and FCNN on ${D}_4$. As can be seen, CNN3 has the worst performance in both datasets and EEGNet has the best performance among all the architectures on ${D}_4$. 

\subsection{Complexity}

\begin{figure}
    \begin{subfigure}[t]{0.5\textwidth}
        \centering
        \includegraphics[width=0.93\textwidth]{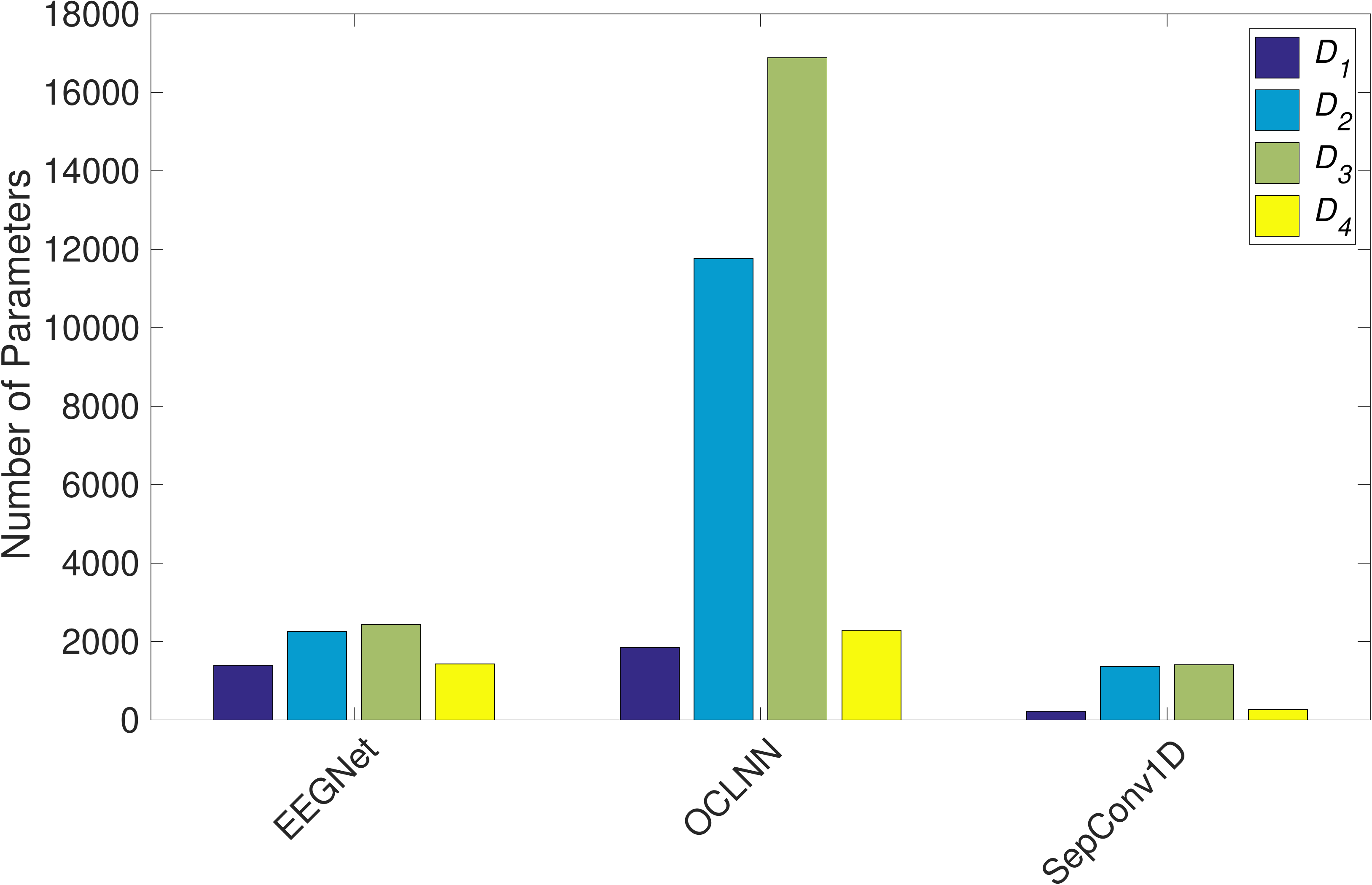}
       \caption{}
    \end{subfigure}
    \begin{subfigure}[t]{0.5\textwidth}
        \centering
        \includegraphics[width=0.82\textwidth]{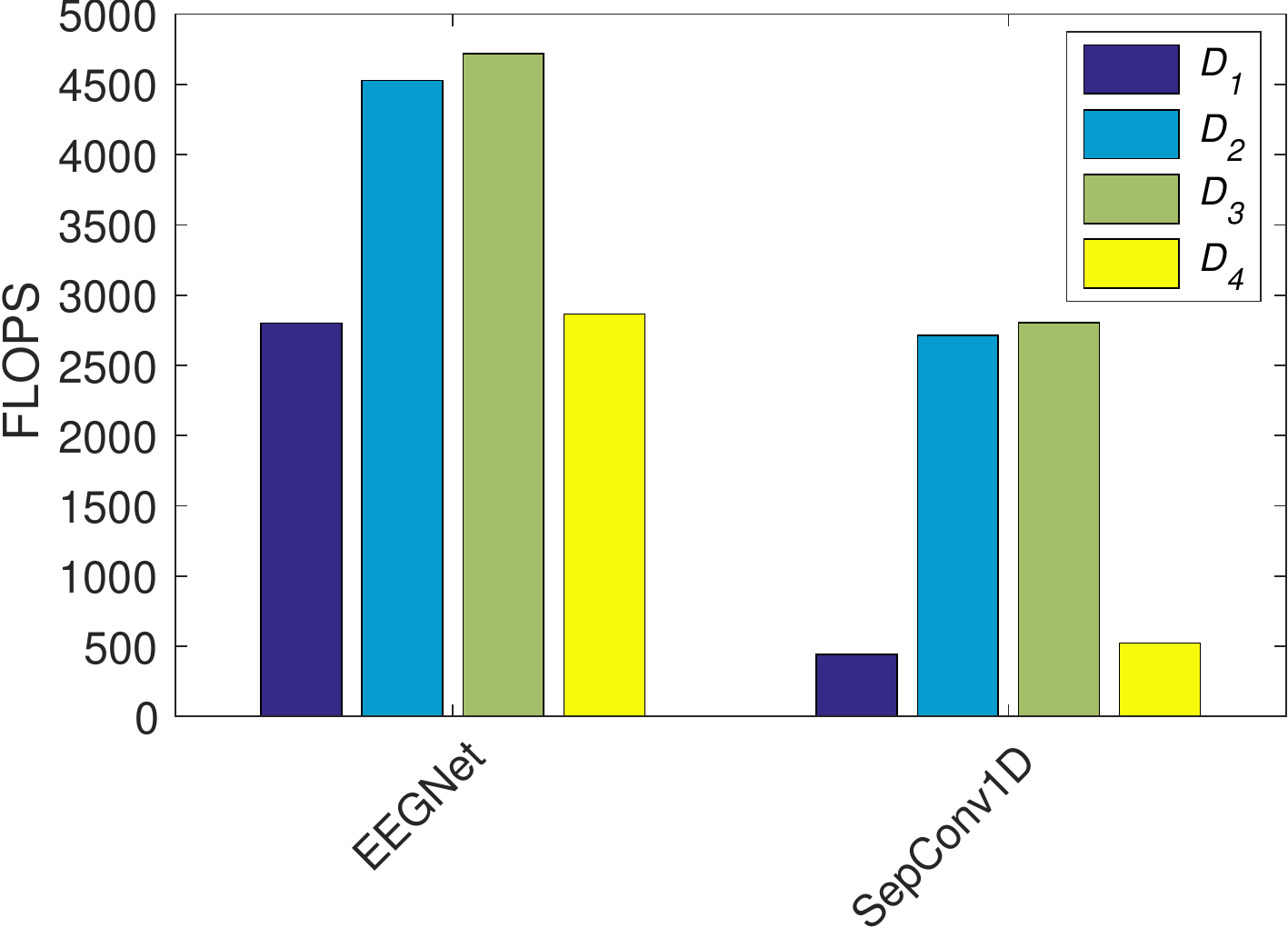}
       \caption{}
    \end{subfigure}
\caption{A comparison of the architectures' complexity for each dataset: (a) trainable parameters and (b) FLOPS = Floating Point Operations per Second. For more details see tables \ref{tab:Parametros} and \ref{tab:Complexity}, respectively.}
\label{fig:parametros}
\end{figure}

Table \ref{tab:Parametros} details the number of trainable parameters used by each architecture for the benchmark datasets. In general, the number of parameters has been decreasing since CNN1 were first introduced for P300 detection in 2011 by Cecotti and Graser \citep{Cecotti2011}. 
It is worth noting that OCLNN, EEGNet, FCNN, and SepConv1D are able to reduce the number of parameters to the order of thousands, at least in one of the datasets (see Figure \ref{fig:parametros}(a)). However, OCLNN requires more than 14,700 parameters for datasets ${D}_2$ and ${D}_3$. Something similar occurs with FCNN, which uses almost 20,000 parameters for ${D}_2$. In contrast, EEGNet manages to keep the number of parameters more or less stable and in the order of thousands. In all cases, SepConv1D requires the least amount of parameters: as few as 225 and 265 parameters for datasets ${D}_1$ and ${D}_4$ respectively and at most 1,361 and 1,405 for datasets ${D}_2$ and ${D}_3$ respectively.

\begin{table}
\centering
\scriptsize
\begin{tabular}{|l|c|c|c|c|c|}
\hline
\textbf{Architecture} & \textbf{${D}_1$} & \textbf{${D}_2$} & \textbf{${D}_3$} & \textbf{${D}_4$} \\
\hline\hline
 CNN-1/UCNN-1 & 1,036,922 & 787,502 &   1,207,502 & 1,036,942\ \\
 CNN-3/UCNN-3 &1,031,009 & 781,067 & 1,201,067 & 1,031,011\ \\
 CNN-R & 19,848,098&  16,445,794 & 21,950,818 & 19,848,290\ \\
 DeepConvNet  & 139,877 &  174,927 & 176,927 & 141,127\ \\
ShallowConvNet &  12,082  & 104,322 & 105,282 & 15,282 \  \\
 BN$^3$  & 44,589 & 39,489 & 47,681 & 44,625\ \\
OCLNN & 1,842 & 11,762 & 16,882 & 2,290\ \\
EEGNet & 1,394 & 2,258 & 2,354 & 1,426 \ \\
FCNN & 2,477 & 19,973 & 30,725 & 3,301  \ \\
SepConv1D & 225 & 1,361 & 1,405 & 265\\
 \hline
\end{tabular}
\caption{Number of trainable parameters of each architecture for all datasets.}
\label{tab:Parametros}
\end{table}

On the other hand, Figure \ref{fig:parametros}(b) takes a closer look at the complexity of EEGNet and SepConv1D in terms of FLOPS. As previously mentioned, the differences between the AUC values of SepConv1D and those of EEGNet are not statistically significant on $D_1$. However, the number of FLOPS and trainable parameters required by SepConv1D is significantly lower compared to EEGNet. 
In contrast, the differences between the AUC values of EEGNet and SepConv1D are statistically significant for within-subject P300 detection on ${D}_2$ and ${D}_4$. Specifically, EEGNet obtained a 0.05 and 0.04 higher mean AUC, but at the cost of 1,814 and 2,342 additional FLOPS and 897 and 1,161 additional parameters for ${D}_2$ and ${D}_4$, respectively. Something similar happens for ${D}_3$, even though the difference between SepConv1D and EEGNet is not statistically significant.
   
\begin{figure}[ht!]
\begin{centering}
    \begin{subfigure}[t]{0.5\textwidth}
        \centering
        \includegraphics[width=6cm,height=4.5cm]{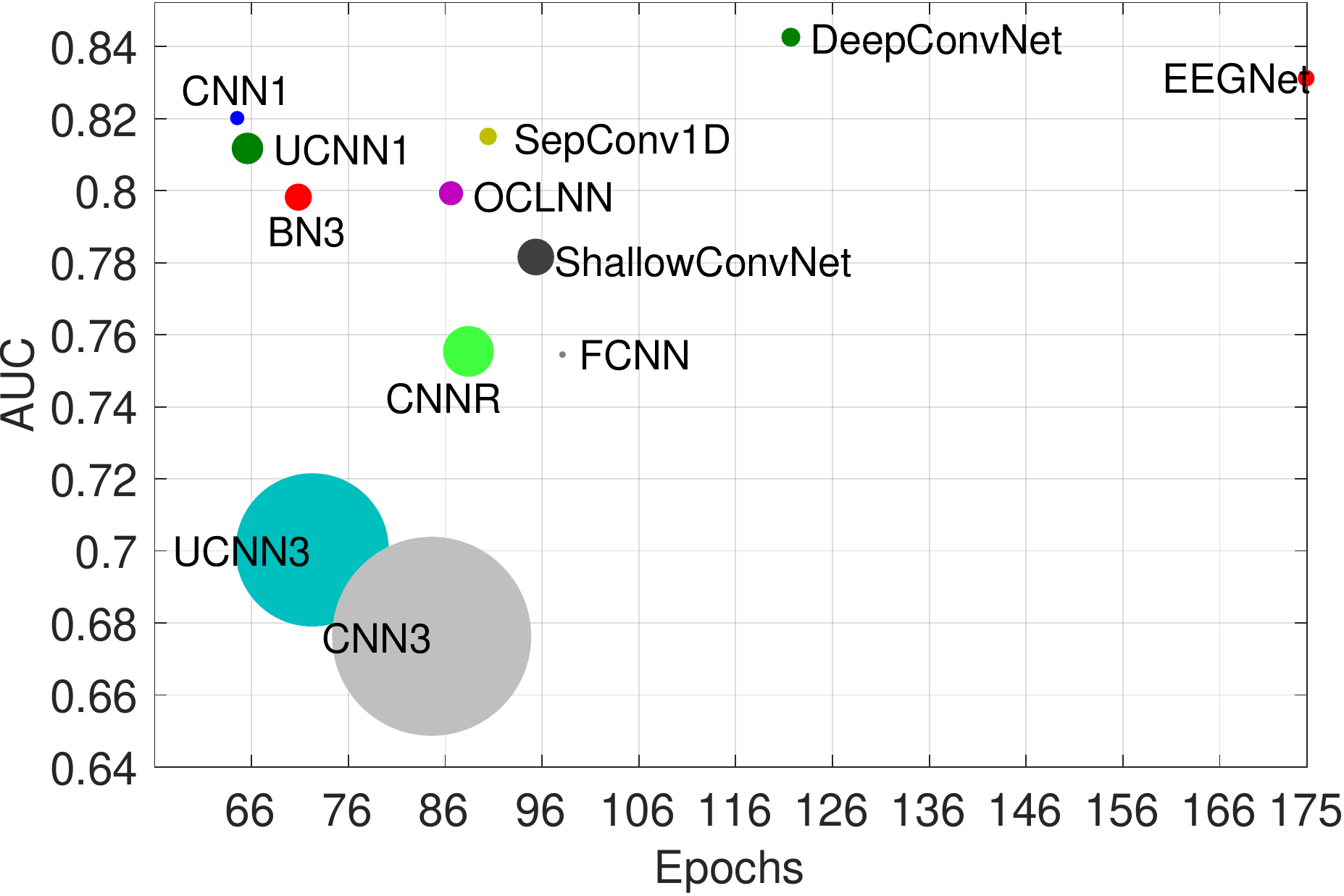}
       \caption{}
    \end{subfigure}%
    ~ 
    \begin{subfigure}[t]{0.5\textwidth}
        \centering
        \includegraphics[width=6cm,height=4.5cm]{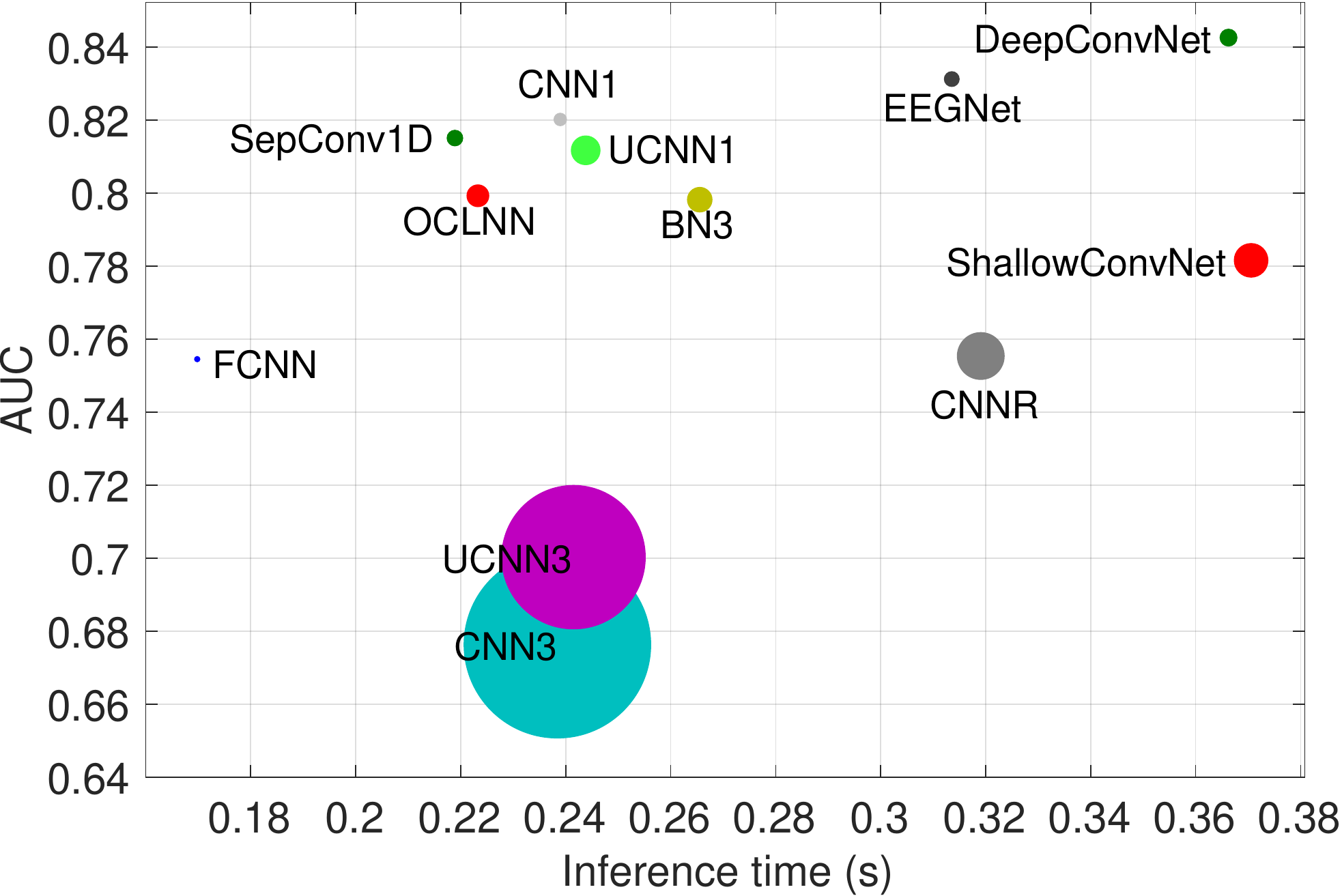}
        \caption{}
    \end{subfigure}

\par\end{centering}
\caption{Comparison between mean AUC values and both (a) epochs  and (b) inference time (see Table \ref{tab:Epochs_within}), obtained by the architectures under analysis for within-subject P300 detection on ${D}_3$. The standard deviations of the AUC values are directly proportional to the diameter of the circles.}
\label{fig:AUC_epochs}
\end{figure}

Finally, we analyzed the training time and the number of epochs needed to meet the early stopping criteria as well as the inference time (i.e., the time needed to process the test set) for the architectures under analysis (see Table \ref{tab:Epochs_within}). To illustrate the performance of the architectures, we compared their mean AUC with both their corresponding epochs (Figure \ref{fig:AUC_epochs}(a)) and inference time (Figure \ref{fig:AUC_epochs}(b)) for within-subject P300 detection on ${D}_3$. The time difference between SepConv1D and the architectures with the highest mean AUC (EEGNet and DeepConvNet) are 0.09 and 0.15 seconds respectively, while the differences in the number of epochs are 84.49 and 31.4 respectively. 

\begin{table}[htbp!]
\centering
\begin{tabular}{|c |c |c | c | c |c |}
\hline
\multicolumn{2}{|c|}{Architecture}& ${D}_1$ & ${D}_2$& ${D}_3$ & ${D}_4$ \\
\hline\hline
\multirow{2}{*}{CNN1}& E& 97$\pm$33 & 68$\pm$ 8 & 65$\pm$ 6 & 71$\pm$14\\
& TT &5.33$\pm$1.74 & 8.08$\pm$ 1.39 & 59.66$\pm$   9.6 & 5.79$\pm$1.29\\
& IT & 0.04$\pm$0.004 & 0.06 $\pm$0.013 & 0.24$\pm$ 0.015 & 0.05$\pm$ 0.006\\
\hline
\multirow{2}{*}{UCNN1}& E& 88$\pm$27 & 74$\pm$26 & 66$\pm$12 & 76$\pm$24\\   
& TT & 4.89$\pm$1.46 &  8.8$\pm$ 3.01 & 63.93$\pm$ 15.76 & 6.27$\pm$1.98\\
& IT & 0.04$\pm$0.004 & 0.06 $\pm$0.013 &0.24$\pm$ 0.014 & 0.05$\pm$ 0.006\\
\hline
\multirow{2}{*}{CNN3}& E&111$\pm$37 & 86$\pm$28 & 85$\pm$29 & 93$\pm$31\\  
& TT &5.25$\pm$1.75 & 9.68$\pm$ 3.28 &  76.1$\pm$ 28.19 & 6.56$\pm$2.13\\
& IT & 0.04$\pm$0.003 & 0.06 $\pm$0.01 & 0.24$\pm$ 0.013& 0.05$\pm$ 0.004\\
\hline
\multirow{2}{*}{UCNN3}& E & 114$\pm$42 & 78$\pm$13 & 72$\pm$16 & 87$\pm$30\\ 
& TT &5.41$\pm$1.97 & 9.01$\pm$ 1.84 & 68.42$\pm$ 20.46 & 5.87$\pm$1.93\\
& IT & 0.04$\pm$0.003 & 0.06 $\pm$0.009 & 0.24$\pm$ 0.011 & 0.05$\pm$ 0.004\\
\hline
\multirow{2}{*}{CNN-R}& E &  61$\pm$ 2 &167$\pm$29 & 89$\pm$29 & 64$\pm$ 2\\
& TT & 12.46$\pm$0.49 &47.25$\pm$ 6.57 &148.38$\pm$ 45.75 &18.68$\pm$0.67\\
& IT & 0.07$\pm$0.01 &0.09 $\pm$0.026  & 0.32$\pm$ 0.026 & 0.09$\pm$ 0.014\\
\hline
\multirow{2}{*}{DeepConvNet}& E &122$\pm$40 & 79$\pm$10 &122$\pm$25 & 106$\pm$24\\
& TT &13.33$\pm$4.15 &26.41$\pm$ 2.88 &276.22$\pm$ 58.13 &18.16$\pm$3.97\\
& IT & 0.11$\pm$0.007 & 0.13 $\pm$0.013 & 0.37$\pm$ 0.013 & 0.12$\pm$ 0.010\\
\hline
\multirow{2}{*}{ShallowConvNet}& E &177$\pm$29 &144$\pm$45 & 95$\pm$40 &157$\pm$33\\
& TT &16.57$\pm$2.69 &63.41$\pm$18.85 &281.23$\pm$116.77 &25.12$\pm$5.29\\
& IT & 0.06$\pm$0.011 & 0.09 $\pm$0.015 & 0.37$\pm$ 0.018 & 0.06$\pm$ 0.010\\
\hline
\multirow{2}{*}{BN$^3$}& E &113$\pm$21 & 77$\pm$ 4 & 71$\pm$ 3 & 95$\pm$ 9\\
& TT & 4.04$\pm$ 0.7 &  8.8$\pm$ 1.31 & 71.08$\pm$ 10.78 & 5.06$\pm$0.56\\
& IT & 0.07$\pm$0.001 & 0.09 $\pm$ 0.007 & 0.27$\pm$ 0.009 & 0.08$\pm$ 0.002\\
\hline
\multirow{2}{*}{EEGNet}& E &200$\pm$ 3 &166$\pm$30 &175$\pm$30 &198$\pm$ 7\\
& TT &17.18$\pm$ 0.5 &53.69$\pm$ 7.28 &360.66$\pm$ 62.81 &27.67$\pm$1.13\\
& IT & 0.08$\pm$0.005 & 0.10 $\pm$ 0.005 & 0.31$\pm$ 0.008 & 0.09$\pm$ 0.005\\
\hline
\multirow{2}{*}{OCLNN}& E &199$\pm$ 5 &129$\pm$41 & 87$\pm$11 &161$\pm$26\\
& TT & 4.55$\pm$0.28 &11.88$\pm$ 2.84 & 75.69$\pm$ 17.95 & 5.87$\pm$0.99\\
& IT & 0.04$\pm$0.002 & 0.05 $\pm$ 0.003 & 0.22$\pm$ 0.005 & 0.04$\pm$ 0.002\\
\hline
\multirow{2}{*}{FCNN}& E &197$\pm$ 7 & 89$\pm$21 & 98$\pm$11 &132$\pm$12\\
& TT & 3.74$\pm$0.21 & 5.71$\pm$ 1.26 & 53.88$\pm$  7.23 & 4.04$\pm$0.38\\
& IT & 0.02$\pm$0.001 & 0.04$\pm$0.002 & 0.17$\pm$ 0.014 & 0.03$\pm$ 0.001\\
\hline
\multirow{2}{*}{SepConv1D}& E &199$\pm$ 5 &104$\pm$14 & 90$\pm$12 &183$\pm$24\\
& TT & 5.34$\pm$0.32 &10.61$\pm$ 2.01 & 80.02$\pm$  14.7 & 8.22$\pm$1.14\\
& IT & 0.03$\pm$0.002 & 0.05 $\pm$0.003 & 0.22$\pm$ 0.009 & 0.04$\pm$ 0.002\\
\hline
\end{tabular}
\caption{Epochs (E), train time (TT), and inference time (IT) obtained by the architectures under analysis for single-trial within-subject P300 detection on each dataset.}
\label{tab:Epochs_within}
\end{table}

\section {Conclusions}\label{sec:Conclusions}
In this paper, we presented SepConv1D, a simple Convolutional Neural Network architecture consisting of a depthwise separable 1D convolutional layer followed by a Sigmoid classification neuron. The proposed architecture can perform within-subject and cross-subject single-trial P300 detection on a par with state-of-the-art CNN-based architectures but with fewer parameters and a lower computational cost. 
Surprisingly, competitive AUC values were achieved with only one filter in some datasets for within-subject detection and with four filters in all cases. 
Increasing the number of filters beyond resulted in small to no improvements in detection performance.
Our results suggest that the P300 component could be effectively detected from EEG signals with a few filters.   
This contrasts with many of the analyzed architectures, which consists of multiple convolutional layers with several filters each.

In general, when the number of channels was larger (i.e. in $D_2$ and $D3$), complexity remain relatively stable for architectures that perform a spatial filtering of the input channels, such as EEGNet, CNN-1 or BN$^3$; in contrast, complexity grew larger for architectures without a spatial filtering, such as OCLNN and SepConv1D. However, some works~\citep{Krusienski2008, Polich2006, Alvarado2016b} have shown that large number of channels are not necessary for P300 detection, so the complexity of these architectures could be controlled by properly selecting the relevant channels in advance, instead of learning combinations of the channels with an additional convolutional layer. 
Remarkably, the performance of CNN-1 was close to those of the best architectures (i.e., EEGNet and DeepConv1D), which suggests that gains in performance have not been as significant since CNN1 was first introduced in 2011, and the most important advances have come in the reduction of the architecture complexity. 

In the four benchmark datasets used in this work, SepConv1D required the lowest number of trainable parameters and FLOPS, and only the FCNN has a lower inference time but also a lower performance. In the dataset with the largest number of subjects, SepConv1D only required 225 parameters, while EEGNet (the architecture with the closest number of parameters) required 1,474, even though their performance was not significantly different. Additionally, the number of FLOPS and inference time required by SepConv1D were significantly lower: EEGNet required 2,801 FLOPS and 0.000274 ms of inference time, while SepConv1D required only 443 FLOPS and less than a half of inference time. 

The performance difference between EEGNet and SepConv1D was statistically significant for within-subject P300 detection on two datasets, where EEGNet obtained a 0.05 and 0.04 higher mean AUC but required 1,814 and 2,342 more FLOPS and 897 and 1,161 additional parameters, and took 0.05 more seconds for inference. Moreover, we analyzed the time and the number of epochs needed for the architectures under analysis to meet the early stopping criteria during training. Compared to the top performing architectures (i.e., EEGNet and DeepConvNet), SepConv1D required less training time. For cross-subject P300 detection, there were no statistically significant differences in performance between the architectures on any of the datasets. 

On the other hand, it is well-known that EEG provides temporal resolution in the millisecond range. This feature is important to BCIs since they can be powerful tools for monitoring cognitive states while performing normal tasks in real-world environments. To  continuously  acquire  and  process  EEG, while allowing near-complete  freedom  of  movement  of  the head  and  body \cite{Lin2009}, it is required to build cheap, light, wearable, and embedded hardware. To that end, low-complexity algorithms need to be implemented. Due to the reduced number of parameters and FLOPS required for its execution, SepConv1D could be adequate to be implemented in these wearable devices.  

Finally, as future work, SepConv1D could potentially be applied to classify other ERPs. For instance, Error-Related Potentials (ErrPs) are also widely used by BCIs \cite{Wilson2019}. ErrPs are evoked potentials elicited when the subject detects an execution error or an outcome error \cite{Spuler2015}. This allows users to feedback the BCI, thus improving their interaction with the system. The ErrPs have both morphological characteristics and a behavior similar to those of the P300 component; although the amplitude of the potential and the time in which it is evoked vary. Because of these similarities, we believe that SepConv1D could also be suitable for detecting ErrPs.

\section {Acknowledgment}
This work was supported by UNAM through the PAPIIT grant IA104016 and by the PRODEP-SEP grant 2017 for the project "Interfaces Brain Computer with perspectives to its application in service robots". We thank the editor and reviewers for the invaluable comments that helped improve this manuscript.

\appendix
\section{Activation Functions}\label{sec:Appendix}
In this appendix, we describe the activation functions used by the analyzed architectures.

\begin{itemize}
 
\item The $Log$ activation function is a logarithmic function  bounded in a range of $[1e-7, 10000]$.
\begin{equation}
f \left(z_{i}^{(l)}\right) = log\left(z_{i}^{(l)}\right). \label{eq:log}
\end{equation}
 
\item The Square activation function does not have a stable range since they explode in magnitude quickly. Since its output is a large value, this function tends to result in bad generalization. Additionally, it takes longer to converge than other activation functions. It is defined as 

\begin{equation}
f \left(z_{i}^{(l)}\right) = z_{i}^{(l)2}.\label{eq:square}
\end{equation}

\item The Sigmoid activation function is a sigmoidal function in the range $[0,1]$. It is defined as 
\begin{equation}
f \left(z_{i}^{(l)}\right) = \frac{1}{1+e^{-z_{i}^{(l)}}}.
\end{equation}

\item The Hyperbolic Tangent activation function ($tanh$) is a sigmoidal function in the range $[-1,1]$. It is defined as 
\begin{equation}
f \left(z_{i}^{(l)}\right) = tanh \left(z_{i}^{(l)}\right) =  
\frac{e^ {2z_{i}^{(l)}}-1}{e^ {2z_{i}^{(l)}}+1}.\label{eq:tanh}
\end{equation}

\item The Scaled Hyperbolic Tangent activation function ($stanh$) \citep{Lecun1989} is a sigmoidal function in the range $[-1,1]$ whose advantage is that the negative inputs will be mapped strongly negative and the zero inputs will be mapped near zero. It is defined as 
\begin{equation}
f \left(z_{i}^{(l)}\right) = 1.7159 tanh \left( \frac{2}{3} z_{i}^{(l)} \right).\label{eq:stanh}
\end{equation}

\item The Softmax activation function \citep{Goodfellow2016} turns scores to probabilities that sum to one. It is defined as 

\begin{equation}
f \left(z_{i}^{(l)}\right) =  \frac{e^ {z_{i}^{(l)}}}{\sum^{j} e^ {z_{j}^{(l)}}}.\label{eq:softmax}
\end{equation}

\item The Rectified Linear Unity (ReLU)  alleviates the vanishing and exploding gradient problems in deep neural networks that are usually associated with saturated activation functions such as Sigmoid or $tanh$. The activation function rectifies the inputs $\mathbf{z} $ by setting the negative values to zero and by keeping the positive values \citep{Nair2010}, which enables faster convergence during training:

\begin{equation}
f \left(z_{i}^{(l)} \right)=max \left(0, z_{i}^{(l)}  \right).
\end{equation}

\item As with ReLU, the Exponential Linear Unit (ELU) \citep{Clevert2015} alleviates the vanishing and exploiting gradient problems via the identity for positive values. It has negative values, thus allowing to push the mean of the activations closer to zero:

\begin{equation}
f \left(z_{i}^{(l)} \right)=
\begin{cases}
  z_{i}^{(l)} &\enskip\text{if } z_{i}^{(l)} > 0 \\
 \phi \left(e^{z_{i}^{(l)}}-1 \right)&\enskip\text{if } z_{i}^{(l)} \leq 0 \\
\end{cases}.
\end{equation}

The ELU hyperparameter $\phi$ controls the value to which an ELU saturates for negative inputs.

\end{itemize}

\section{Architectures}\label{sec:AppendixB}
This appendix provides the specifications of the state-of-the-art CNN architectures analyzed in this paper. For the sake of simplicity, the reported number of filters, input size, number of parameters, and output size of every layer correspond only to those obtained for the dataset $D_1$.

\setcounter{table}{0}
\begin{table}[ht]
\centering
\begin{threeparttable}
\scriptsize
\begin{tabular}{|l|p{0.7cm}|p{0.7cm}|p{1.3cm}|p{1.2cm}|p{1.4cm}|p{3cm}|}
\hline
\textbf{Layer} &  \textbf{No. filters} & \textbf{Size} & \textbf{No. params} & \textbf{Output}  &\textbf{Activation function} & \textbf{Options} \\
\hline\hline
Input & & $ 206 \times 6$ & & & &\\
Conv1D & $10$ & $1$ & 70 & $(206,10)$ & & padding = same, data format = "Channels last", bias and kernel initializer =  ** \\ 
Activation & & & & $(206,10)$ & $stanh$ &  \\
& & & & & & \\
Conv1D & $50$ & 13  & $6,550$ & $(206,50)$& & padding = same, data format = "Channels first", bias and kernel initializer =  **  \\ 
Activation & & & & $(206,50)$ & $stanh$ &  \\
& & & & & & \\
Flatten	& & & &$(10,300)$ & &\\
Dense & $100$ & &$1,030,100$ & $(100)$&  Sigmoid &  \\
& & & & & & \\
Dense & $2$ & & $202$ & $(2)$ & Sigmoid$^1$/ Softmax$^2$ & \\\hline
  \end{tabular}
\end{threeparttable}
\caption{CNN-1 and UCNN-1 architectures. $^1$ CNN-1 uses Sigmoid activation. $^2$ UCNN-1 uses Softmax activation. }
\label{tab:CNN-1_Arq}
\end{table}

\begin{table}[ht]
\centering
\begin{threeparttable}
\scriptsize
\begin{tabular}{|l|p{0.7cm}|p{0.7cm}|p{1.3cm}|p{1.2cm}|p{1.4cm}|p{3cm}|}
\hline
\textbf{Layer} &  \textbf{No. filters} & \textbf{Size} & \textbf{No. params} & \textbf{Output}  &\textbf{Activation function} & \textbf{Options} \\
\hline\hline
Input & & $ 206 \times 6$ & & & &\\
Conv1D & $1$ & $1$ & 7 & $(206,1)$ & & padding = same, data format = "Channels last", bias and kernel initializer =  ** \\ 
Activation & & & & $(206,1)$ & $stanh$ &  \\
& & & & & & \\
Conv1D & $50$ & 13  & $700$ & $(206,50)$& & padding = same, data format = "Channels last", bias and kernel initializer =  **  \\ 
Activation & & & & $(206,50)$ & $stanh$ &  \\
& & & & & & \\
Flatten	& & & &$(10,300)$ & &\\
Dense & $100$ & &$1,030,100$ &$(100)$ &  Sigmoid &  \\
& & & & & & \\
Dense & $2$ & & $202$ & $(2)$ & Sigmoid$^1$/ Softmax$^2$ & \\\hline
  \end{tabular}
\end{threeparttable}
\caption{CNN-3 and UCNN-3 architectures. $^1$ CNN-3 uses Sigmoid activation. $^2$ UCNN-3 uses Softmax activation. }
\label{tab:CNN-3_Arq}
\end{table}

\begin{table}[ht]
\centering
\begin{threeparttable}
\scriptsize
\begin{tabular}{|l|p{0.7cm}|p{0.7cm}|p{1.3cm}|p{1.2cm}|p{1.4cm}|p{2cm}|}
\hline
\textbf{Layer} &  \textbf{No. filters} & \textbf{Size} & \textbf{No. params} & \textbf{Output}  &\textbf{Activation function} & \textbf{Options} \\
\hline\hline
Input & & $ 206 \times 6$ & & & &\\
Conv1D & $96$ & $1$ & 672 & $(206,96)$ & & activity regularizer$^1$ \\ 
Activation & & & & $(206,96)$ & ReLU &  \\
Max Pooling 1D &3 & stride 2& &$(102,96)$ & & \\
& & & & & & \\
Conv1D & 128 & 6  & 73,856 & $(97,128)$& & \\ 
Activation & & & & $(97,128)$ & ReLU &  \\
Max Pooling 1D & 3& stride 2 & &$(48,128)$ & & \\
& & & & & & \\
Conv1D & 128 & 6  & 98,432 & $(43,128)$& &  \\ 
Activation & & & & $(43,128)$ & ReLU &  \\
& & & & & & \\
Flatten	& & & &$(5,504)$ & &\\
Dense & 2,048 & &$11,274,240$ &$(2,048)$ &  ReLU &  \\
Dropout & & & &$(2,048)$ &  & $p=0.8$ \\ 
Dense & 4,096 & & 8,392,704 & $(4,096)$ & ReLU & \\
Dropout & & & &$(4,096)$ &  & $p=0.8$ \\
Dense & 2 & & 8,194 & $(2)$ &  & \\
Activation & & & & $(2)$ & Softmax &  \\\hline
  \end{tabular}
\end{threeparttable}
\caption{CNN-R architecture. $^1$ Activity regularizer $=0.01 \sum_{t=1}^{N-1} (a^{(l)}_{t+1}-a^{(l)}_{t})^{2}$, where $a^{(l)}_{t}$ is the output of the convolutional layer $l$ at time $t$.}
\label{tab:CNNR_Arq}
\end{table}

\clearpage
\begin{table}[htb]
\centering
\scriptsize
\begin{tabular}{|l|p{0.7cm}|p{0.7cm}|p{1.3cm}|p{1.2cm}|p{1.4cm}|p{3cm}|}
\hline
\textbf{Layer} &  \textbf{No. filters} & \textbf{Size} & \textbf{No. params} & \textbf{Output}  &\textbf{Activation function} & \textbf{Options} \\
\hline\hline
Input & & $6 \times 206$ & & & & \\
Reshape & &$1 \times 6 \times 206$ & & & & \\
Conv2D & 25 & $1 \times 5$ & $150$ & $(25,6,202)$ & & max norm $=2$ \\ 
Conv2D & 25 & $6 \times 1$ & $40,025$ & $(25,1,202)$ & & max norm $=2$ \\ 
& & & & & &\\
BatchNorm & & &$100$ &$(25,1,202)$ & & data format = "Channels first", $\epsilon= 1\times10^{-05} $, momentum $= 0.1$\\
Activation & & & & $(25,1,202)$ & ELU &  \\
MaxPool2D & & $1 \times 2$ pool size and stride   & & $(25,1,101)$ & &\\
Dropout & & & & $(25,1,101)$ & &$p=0.5$\\  
Conv2D	& 50 & $1 \times 5$ & 6,300 & $(50,1,97)$ & & max norm $=2$ \\
& & & & & &\\
BatchNorm & & &$200$ &$(50,1,97)$ & & data format = "Channels first", $\epsilon= 1\times10^{-05} $, momentum $= 0.1$\\
Activation & & & & $(50,1,97)$ & ELU &  \\
MaxPool2D & & $1 \times 2$ pool size and stride   & & $(50,1,48)$ & &\\
Dropout & & & & $(50,1,48)$ & &$p=0.5$\\
Conv2D & 100 & $1 \times 5$ & $25,100$ & $(100,1,44)$ & & max norm $=2$ \\
& & & & & &\\
BatchNorm & & &$400$ &$(100,1,44)$ & & data format = "Channels first", $\epsilon= 1\times10^{-05} $, momentum $= 0.1$\\
Activation & & & & $(100,1,44)$ & ELU &  \\
MaxPool2D & & $1 \times 2$ pool size and stride   & & $(100,1,22)$ & &\\
Dropout & & & & $(100,1,22)$ & &$p=0.5$\\
Conv2D & 200 & $1 \times 5$ & $100,200$ & $(200,1,18)$ & & max norm $=2$ \\
& & & & & &\\
BatchNorm & & &$800$ &$(200,1,18)$ & & data format = "Channels first", $\epsilon= 1\times10^{-05} $, momentum $= 0.1$\\
Activation & & & & $(200,1,18)$ & ELU &  \\
MaxPool2D & & $1 \times 2$ pool size and stride   & & $(200,1,9)$ & &\\ 
Dropout & & & & $(200,1,9)$ & &$p=0.5$\\
& & & & & &\\
Flatten	& & & & $(1,800)$ & &\\
& & & & & &\\
Dense & $2$ & & $4,802$ &(2) & Softmax & max norm = 0.5\\\hline
\end{tabular}
\caption{DeepConvNet architecture. Version modified by \citep{Lawhern2018}.}
\label{tab:DeepConvNet_Arq}
\end{table}

\begin{table}[h!]
\centering
\scriptsize
\begin{tabular}{|l|p{0.7cm}|p{0.7cm}|p{1.3cm}|p{1.2cm}|p{1.4cm}|p{3cm}|}
\hline
\textbf{Layer} &  \textbf{No. filters} & \textbf{Size} & \textbf{No. params} & \textbf{Output}  &\textbf{Activation function} & \textbf{Options} \\
\hline\hline
Input & & $6 \times 206$ & & & &   \\
Reshape & &$1 \times 6 \times 206$& & && \\
Conv2D & 40 & $1 \times 13$ & 560 &$(40,6,194)$& & max norm constraint function $=2$ \\
Conv2D & 40 & $6 \times 1 $ & 9,600 & $(40,1,194)$& & max norm constraint function $=2$, doesn't use bias \\
BatchNorm  & & & $160$ & $(40,1,194)$ & & axis to be normalized =1, $\epsilon= 1\times10^{-05} $, momentum $= 0.1$\\
Activation & & & & $(40,1,194)$ & Square & \\
AveragePool2D & & pool $1 \times 35$, stride $1 \times 7$  & & $(40,1,23)$ & &\\
Activation & & & & $(40,1,23)$ & $Log$ &\\
Dropout & & & &$(40,1,23)$ & & $p=0.5$\\
Flatten	 & & & & $(900)$ & &\\
  & & & & & &\\
Dense & 2  & & $1,842$ & $(2)$ & Softmax & max norm constraint function $= 0.5$ \\\hline
\end{tabular}
\caption{ShallowConvNet architecture. Version modified by  \citep{Lawhern2018}.}
\label{tab:ShallowConvNet_Arq}
\end{table}

\clearpage
\begin{table}[htb]
\centering
\scriptsize
\begin{tabular}{|l|p{0.7cm}|p{0.7cm}|p{1.3cm}|p{1.2cm}|p{1.4cm}|p{3cm}|}
\hline
\textbf{Layer} &  \textbf{No. filters} & \textbf{Size} & \textbf{No. params} & \textbf{Output}&\textbf{Activation function}  & \textbf{Options} \  \\
\hline\hline
Input & & $206 \times 6$  &  & & &  \\
BatchNorm  & & & 24 & $206 \times 6$ &  & \\
Conv1D & $16$ & $1$ & 112 & $206 \times 16$ &  ReLU & bias initializer = Glorot Uniform \\
Conv1D & $16$ & $20$ & 5,136 & $11 \times 16$ &  ReLU & bias initializer = Glorot Uniform, strides = 20, padding = same\\
BatchNorm  & &  & 64 &$11 \times 16$ &  & \\
Activation  & & & & $11 \times 16$ & ReLU & \\
& & & & & &\\
Flatten	& & & & 176 & &\\
Dense & & $128$ & 22,656 & 128 &  $tanh$ &bias initializer = Glorot Uniform \\
Dropout  & &  & & 128 & & $p=0.8$ \\
Dense & & $128$ & 16,512 & 128 &  $tanh$ & bias initializer = Glorot Uniform \\
Dropout & & & & 128 & & $p=0.8$\\
Dense & 1 & &  129 & 1 & Sigmoid & bias initializer = Glorot Uniform \\\hline
\end{tabular}
\caption{BN$^3$ architecture.}
\label{tab:BN3_Arq}
\end{table}

\begin{table}[ht]
\centering
\scriptsize
\begin{tabular}{|l|p {0.8cm}|p{1cm}|p{1cm}|p{1.5cm}|p{1.5cm}|p{3.5cm}|}
\hline
\textbf{Layer} &  \textbf{No. filters} & \textbf{Size} & \textbf{No. params} & \textbf{Output}&\textbf{Activation function}  & \textbf{Options} \  \\
\hline\hline
Input & & $6 \times 206$ & &  & &  \\
Reshape & & & & $1 \times 6 \times 206$& & \\
Conv2D & 8  & $1\times64$&$512$ & $(4,6,206) $  & Linear & padding $=$ same, doesn't use bias \\ 
BatchNorm  & & & $16$ & $(4,6,206)$ & &\\
DepthwiseConv2D & &$6 \times 1$ & $48$ & $(8,1,206)$ & Linear & doesn't use bias, number of depthwise convolution output channels $=2$, max norm constraint function $=1$\\
& & & & & &\\
BatchNorm  & & & $32$ & $(8,1,206)$ & &\\
Activation & & & &$(8,1,206)$ & ELU &  \\
AveragePool2D & & $1\times4$ & & $(8,1,51)$ & &\\ 
Dropout & & & & $(8,1,51)$ & &$p$ \\  
& & & & & &\\
SeparableConv2D & 16& $1\times16$ & $192$ &  $(8,1,51)$ & Linear & padding $=$ same, doesn't use bias \\
BatchNorm  & & & $32$ & $(8,1,51)$ & &\\
Activation & & & &$(8,1,51)$ &ELU &  \\
AveragePool2D & & $1\times8$ & & $(8,1,6)$  & &\\ 
Dropout & & & & $(8,1,6)$ & &$p$ \\  
& & & & & &\\
Flatten	& & & &$(48)$ & &\\
& & & & & &\\
Dense  & $2$& &$98$ & $(2)$ & Softmax  & max norm constraint regularization = 0.25\\\hline
\end{tabular}
\caption{EEGNet architecture, where $p=0.25$ or $0.5$ (for cross-subject or within-subject classification, respectively). Adapted from \citep{Lawhern2018}.}
\label{tab:EEGNet_Arq}
\end{table}

\begin{table}[htbp!]
\centering
\begin{threeparttable}
\scriptsize
\begin{tabular}{|l|p{0.7cm}|p{1.5cm}|p{1.5cm}|p{1.5cm}|p{1.4cm}|p{3cm}|}
\hline
\textbf{Layer} &  \textbf{No. filters} & \textbf{Size} & \textbf{No. params} & \textbf{Output}  &\textbf{Activation function} & \textbf{Options} \\
\hline\hline
Input & & $ 206 \times 6$ & & & &\\
ZeroPadding1D &  &  &  & $(210,6)$ & & padding = 2 \\ 
Conv1D & $16$ & kernel and stride = 14 & 1,360 & $(15,16)$ & & kernel and bias regularizer use L2 $= 0.01$ \\ 
Activation & & & & $(15,16)$ & ReLU &  \\
Dropout & & & &$(15,16)$ &  & $p=0.25$ \\ 
Flatten	& & & &$(240)$ & &\\
Dense & 2 & & 482 & $(2)$ &  & \\
Activation & & & & $(2)$ & Softmax &  \\\hline
  \end{tabular}
\end{threeparttable}
\caption{OCLNN architecture.}
\label{tab:OCLNN_Arq}
\end{table}

\begin{table} [ht!]
\centering
\scriptsize
\begin{tabular}{|l|p {1.2cm}|p{1.3cm}|p{1cm}|p{2.2cm}|}
\hline
\textbf{Layer} &\textbf{No. filters} & \textbf{No. params} & \textbf{Output}&\textbf{Activation function}   \  \\
\hline\hline
Input & & & $6 \times 206$ & \\
Reshape & & & $1236\times 1 $& \\
Dense  & $2$&$2,474$ & $(2)$ & $tanh$ \\
Flatten	& & & $(2)$ &\\
& & & & \\
Dense  & $1$&$3$ & $(1)$ & Sigmoid \\\hline
\end{tabular}
\caption{FCNN architecture.}
\label{tab:FCNN_Arq}
\end{table}

\section{Per-subject mean AUC values and architecture complexities}\label{sec:AppendixC}

The results from tables \ref{tab:AUCdetail-within-lini}-\ref{tab:AUCdetail-within-als} are rounded up to four decimal.
\setcounter{table}{0}
\begin{sidewaystable}[hbt!]
\centering
\caption{Per-subject mean AUC obtained by the architectures under analysis for within-subject P300 detection on dataset ${D}_1$.}
\begin{tabular}{|p{1cm}|p{0.8cm}|p{1cm}|p{0.9cm}|p{1.2cm}|p{1.2cm}|p{1.5cm}|p{1.6cm}|p{1cm}|p{1.1cm}|p{1.1cm}|p{0.9cm}|p{1.8cm}|}
\hline
\textbf{Subject} &
\textbf{CNN1} &  \textbf{UCNN1} & \textbf{CNN3} & \textbf{UCNN3} & \textbf{CNN-R}  &\textbf{Deep ConvNet} & \textbf{Shallow ConvNet} &  \textbf{BN$^3$} & \textbf{EEGNet} & \textbf{OCLNN} & \textbf{FCNN}  &\textbf{SepConv1D}\\
\hline\hline

ACS& 0.9123	& 0.9165	& 0.6894	& 0.7288	& 0.8881	& 0.9118	& 0.8249	& 0.8981	& 0.8974	& 0.9113	& 0.919	& 0.9129\\
APM& 0.9314	& 0.9347	& 0.7721	& 0.7919	& 0.9177	& 0.9398	& 0.899	& 0.926	& 0.9309	& 0.936	& 0.9327	& 0.93\\
ASG& 0.908	& 0.9108	& 0.771	& 0.7795	& 0.9118	& 0.9218	& 0.8969	& 0.9087	& 0.9222	& 0.9139	& 0.9073	& 0.9149\\
ASR& 0.7311	& 0.729	& 0.6263	& 0.6397	& 0.7147	& 0.7423	& 0.6675	& 0.7316	& 0.7421	& 0.7367	& 0.7311	& 0.7302\\
CLL& 0.8479	& 0.8572	& 0.6453	& 0.6616	& 0.8126	& 0.8448	& 0.7267	& 0.834	& 0.8309	& 0.8441	& 0.8583	& 0.8478\\
DCM& 0.9421	& 0.9458	& 0.8187	& 0.8621	& 0.9353	& 0.9503	& 0.9096	& 0.9442	& 0.9404	& 0.9467	& 0.943	& 0.9434\\
DLP& 0.8042	& 0.8206	& 0.6171	& 0.6655	& 0.7903	& 0.83	& 0.7324	& 0.791	& 0.8224	& 0.8153	& 0.814	& 0.815\\
DMA& 0.8516	& 0.8667	& 0.6755	& 0.709	& 0.8397	& 0.8655	& 0.7499	& 0.8474	& 0.8373	& 0.857	& 0.8641	& 0.8599\\
ELC& 0.9077	& 0.9086	& 0.7526	& 0.813	& 0.8939	& 0.9164	& 0.8548	& 0.9008	& 0.9022	& 0.9064	& 0.9091	& 0.9105\\
FSZ& 0.9035	& 0.9057	& 0.7347	& 0.7724	& 0.8896	& 0.9116	& 0.8602	& 0.8977	& 0.9065	& 0.9046	& 0.9062	& 0.9081\\
GCE& 0.8674	& 0.8745	& 0.6963	& 0.7533	& 0.852	& 0.8742	& 0.8223	& 0.8659	& 0.8673	& 0.8757	& 0.8697	& 0.8696\\
ICE& 0.8882	& 0.893	& 0.7108	& 0.7503	& 0.8776	& 0.9013	& 0.7757	& 0.8852	& 0.8763	& 0.8916	& 0.8931	& 0.8903\\
JCR& 0.8577	& 0.8713	& 0.7078	& 0.7582	& 0.8505	& 0.8746	& 0.8038	& 0.8537	& 0.8623	& 0.8667	& 0.8672	& 0.8666\\
JLD& 0.8931	& 0.8931	& 0.7383	& 0.8142	& 0.8877	& 0.9051	& 0.8582	& 0.891	& 0.8968	& 0.896	& 0.8938	& 0.8929\\
JLP& 0.8773	& 0.8871	& 0.6724	& 0.7177	& 0.8745	& 0.899	& 0.8532	& 0.8817	& 0.8864	& 0.8821	& 0.8842	& 0.8878\\
JMR& 0.9187	& 0.9244	& 0.7602	& 0.8377	& 0.9061	& 0.925	& 0.8797	& 0.9117	& 0.9151	& 0.9185	& 0.929	& 0.9211\\
JSC& 0.8223	& 0.837	& 0.622	& 0.67	& 0.8081	& 0.8483	& 0.748	& 0.8233	& 0.8317	& 0.8379	& 0.8388	& 0.8361\\
JST& 0.9675	& 0.9684	& 0.8433	& 0.9121	& 0.9635	& 0.9715	& 0.951	& 0.9641	& 0.9699	& 0.9681	& 0.9675	& 0.9677\\
LAC& 0.9379	& 0.9485	& 0.8037	& 0.8599	& 0.9457	& 0.9524	& 0.9322	& 0.9413	& 0.9438	& 0.9497	& 0.9468	& 0.9427\\
LAG& 0.933	& 0.9353	& 0.6993	& 0.8393	& 0.9289	& 0.9435	& 0.9158	& 0.9382	& 0.9412	& 0.9406	& 0.9323	& 0.9404\\
LGP& 0.9288	& 0.9338	& 0.8228	& 0.8587	& 0.9268	& 0.9399	& 0.8939	& 0.9287	& 0.9339	& 0.934	& 0.9348	& 0.9354\\
LPS& 0.863	& 0.8668	& 0.6406	& 0.764	& 0.8473	& 0.871	& 0.802	& 0.8617	& 0.8675	& 0.867	& 0.8666	& 0.8636
\\\hline
  \end{tabular}

\label{tab:AUCdetail-within-lini}
\end{sidewaystable}

\begin{sidewaystable}[htbp!]
\centering
\caption{Per-subject mean AUC obtained by the architectures under analysis for cross-subject P300 detection on dataset ${D}_1$.}
\begin{tabular}{|p{1cm}|p{0.8cm}|p{1cm}|p{0.9cm}|p{1.2cm}|p{1.2cm}|p{1.5cm}|p{1.6cm}|p{1cm}|p{1.1cm}|p{1.1cm}|p{0.9cm}|p{1.7cm}|}
\hline
\textbf{Subject} &
\textbf{CNN1} &  \textbf{UCNN1} & \textbf{CNN3} & \textbf{UCNN3} & \textbf{CNN-R}  & \textbf{Deep ConvNet} & \textbf{Shallow ConvNet} &  \textbf{BN$^3$} & \textbf{EEGNet} & \textbf{OCLNN} & \textbf{FCNN}  &\textbf{SepConv1D}\\
\hline\hline

ACS&0.8527	&0.8653	&0.8392	&0.8366	&0.8545	&0.8695	&0.8151	&0.8725	&0.8685	&0.8539	&0.8554	&0.8727\\
APM&0.8453	&0.8615	&0.3495	&0.8442	&0.8468	&0.8569	&0.8508	&0.8652	&0.8706	&0.8552	&0.8515	&0.8627\\
ASG&0.8680	&0.8914	&0.8680	&0.8878	&0.8892	&0.8995	&0.8753	&0.8813	&0.8972	&0.8741	&0.8712	&0.8917\\
ASR&0.7260	&0.7240	&0.7167	&0.7146	&0.7293	&0.7237	&0.7253	&0.7268	&0.7332	&0.7244	&0.7205	&0.7247\\
CLL&0.7009	&0.7159	&0.7054	&0.6977	&0.7036	&0.7222	&0.7071	&0.7258	&0.7195	&0.7046	&0.7235	&0.7096\\
DCM&0.8332	&0.8960	&0.8421	&0.8903	&0.8819	&0.8901	&0.8840	&0.8860	&0.8899	&0.8900	&0.8851	&0.8806\\
DLP&0.7687	&0.7998	&0.7726	&0.7742	&0.7907	&0.7856	&0.7615	&0.7984	&0.7969	&0.7709	&0.7888	&0.7939\\
DMA&0.7542	&0.8119	&0.7923	&0.7962	&0.7791	&0.7963	&0.7686	&0.8090	&0.7994	&0.7741	&0.7863	&0.7998\\
ELC&0.8433	&0.8738	&0.8477	&0.8613	&0.8758	&0.8600	&0.8608	&0.8818	&0.8747	&0.8809	&0.8774	&0.8798\\
FSZ&0.8204	&0.8563	&0.8315	&0.8496	&0.8392	&0.8506	&0.8480	&0.8625	&0.8590	&0.8448	&0.8488	&0.8579\\
GCE&0.8018	&0.8312	&0.7782	&0.7932	&0.8206	&0.8389	&0.8435	&0.8317	&0.8379	&0.8213	&0.8074	&0.8193\\
ICE&0.7664	&0.7880	&0.7577	&0.7660	&0.7816	&0.7808	&0.7538	&0.7810	&0.7907	&0.7966	&0.7956	&0.8044\\
JCR&0.8350	&0.8308	&0.4988	&0.8241	&0.8458	&0.8174	&0.8117	&0.8390	&0.8213	&0.8296	&0.8266	&0.8274\\
JLD&0.8650	&0.8666	&0.8326	&0.8415	&0.8664	&0.8749	&0.8568	&0.8656	&0.8715	&0.8599	&0.8491	&0.8632\\
JLP&0.8079	&0.8181	&0.8014	&0.8061	&0.8333	&0.8024	&0.8081	&0.8197	&0.8065	&0.8191	&0.7907	&0.8070\\
JMR&0.8007	&0.8523	&0.8374	&0.8378	&0.8192	&0.8479	&0.8427	&0.8337	&0.8419	&0.8451	&0.8277	&0.8406\\
JSC&0.7306	&0.7360	&0.7221	&0.7119	&0.7511	&0.7437	&0.6995	&0.7291	&0.7286	&0.7265	&0.7338	&0.7326\\
JST&0.9125	&0.9194	&0.9082	&0.9154	&0.9105	&0.9148	&0.9127	&0.9052	&0.9251	&0.9115	&0.9101	&0.9111\\
LAC&0.8395	&0.9240	&0.7672	&0.8940	&0.9211	&0.9240	&0.9194	&0.9178	&0.9245	&0.9025	&0.9169	&0.9161\\
LAG&0.8217	&0.8722	&0.8571	&0.8737	&0.8666	&0.8623	&0.9093	&0.8737	&0.8960	&0.8841	&0.8606	&0.8775\\
LGP&0.8704	&0.8951	&0.8752	&0.8937	&0.8930	&0.9022	&0.8765	&0.8994	&0.8966	&0.8952	&0.8913	&0.9006\\
LPS&0.7923	&0.8339	&0.8141	&0.8157	&0.8030	&0.8080	&0.8152	&0.8199	&0.8257	&0.8325	&0.8048	&0.8204		
\\\hline
  \end{tabular}

\label{tab:AUCdetail-cross-lini}
\end{sidewaystable}

\begin{sidewaystable}[htbp!]
\centering
\caption{Per-subject mean AUC obtained by the architectures under analysis for within-subject P300 detection on datasets $D_2$ and $D_3$.}
\begin{tabular}{|p{1.2cm}|p{0.8cm}|p{1cm}|p{0.9cm}|p{1.2cm}|p{1.2cm}|p{1.5cm}|p{1.6cm}|p{1cm}|p{1.1cm}|p{1.1cm}|p{0.9cm}|p{1.8cm}|}
\hline
\multicolumn{13}{|c|}{(a) Dataset  ${D}_2$} \\
\hline\textbf{Subject} &
\textbf{CNN1} &  \textbf{UCNN1} & \textbf{CNN3} & \textbf{UCNN3} & \textbf{CNN-R}  &\textbf{Deep ConvNet} & \textbf{Shallow ConvNet} &  \textbf{BN$^3$} & \textbf{EEGNet} & \textbf{OCLNN} & \textbf{FCNN}  &\textbf{SepConv1D}\\
\hline\hline
Subject1 & 0.9162 & 0.897 & 0.7952 & 0.8458 & 0.9025 & 0.9374 & 0.8871 & 0.8522 & 0.9306 & 0.8995 & 0.8285 & 0.8933\\
Subject2 & 0.847 & 0.842 & 0.7394 & 0.769 & 0.8058 & 0.869 & 0.7477 & 0.7661 & 0.8609 & 0.7970 & 0.7372 & 0.8116
\\\hline
\multicolumn{13}{c}{}\\
\hline
\multicolumn{13}{|c|}{(b) Dataset  ${D}_3$} \\
\hline
\textbf{Subject} & 
\textbf{CNN1} &  \textbf{UCNN1} & \textbf{CNN3} & \textbf{UCNN3} & \textbf{CNN-R}  &\textbf{Deep ConvNet} & \textbf{Shallow ConvNet} &  \textbf{BN$^3$} & \textbf{EEGNet} & \textbf{OCLNN} & \textbf{FCNN}  &\textbf{SepConv1D}\\
\hline
Subject1 & 0.7827 & 0.7657 & 0.6378 & 0.7137 & 0.6974 & 0.7980 & 0.7318 & 0.7532 & 0.7898 & 0.7527 & 0.7197 & 0.7725\\
Subject2 & 0.8590 & 0.8509 & 0.6875 & 0.7153 & 0.8123 & 0.8852 & 0.8308 & 0.8444 & 0.8727 & 0.8436 & 0.7890 & 0.8589
\\\hline
\end{tabular}
\label{tab:AUCdetail-with-BCI}
\end{sidewaystable}

\begin{sidewaystable}[htbp!]
\centering
\caption{Per-subject mean AUC obtained by the architectures under analysis on dataset ${D}_4$}
\begin{tabular}{|p{1cm}|p{0.8cm}|p{1cm}|p{0.9cm}|p{1.2cm}|p{1.2cm}|p{1.5cm}|p{1.6cm}|p{1cm}|p{1.1cm}|p{1.1cm}|p{0.9cm}|p{1.6cm}|}
\hline
\multicolumn{13}{|c|}{(a) Within-subject P300 detection} \\
\hline
\textbf{Subject} &
\textbf{CNN1} &  \textbf{UCNN1} & \textbf{CNN3} & \textbf{UCNN3} & \textbf{CNN-R}  &\textbf{Deep ConvNet} & \textbf{Shallow ConvNet} &  \textbf{BN$^3$} & \textbf{EEGNet} & \textbf{OCLNN} & \textbf{FCNN}  &\textbf{SepConv1D}\\

\hline
s0 &  0.8166 & 	0.8014 & 	0.5950 & 0.6358 & 0.8194 & 0.8487 	& 0.8102 	& 0.7850   & 0.8577 & 0.8066 & 0.8147 & 0.8266 \\
s1 &  0.8092 & 	0.8040 & 	0.6207 & 0.6477 & 0.8138 & 0.8475 	& 0.8163  	& 0.7830 	& 0.8586 & 0.8048 & 0.7962 & 0.8233 \\
s2 &  0.8829 & 	0.8723 & 	0.6984 & 0.7559 & 0.8765 & 0.9005 	& 0.8734 	& 0.8742	& 0.8953 & 0.8772 & 0.8645 & 0.888 \\
s3 & 	 0.8062 & 0.7954 & 	0.6666 & 0.6731 & 0.7987 & 0.8337 	& 0.8059	& 0.8051	& 0.8332 & 0.8020 & 0.7815 & 0.8093 \\
s4 & 	 0.8514 & 0.8464 & 	0.6364 & 0.6755 & 0.8583 & 0.8821 	& 0.8541	& 0.8373	& 0.8800 & 0.8616 & 0.8326 & 0.8611 \\
s5 &  0.8759 & 	0.8736 & 	0.6449 & 0.7074 & 0.8636 & 0.8917  & 0.8644	& 0.8572	& 0.8940 & 0.8730 & 0.8640 & 0.8798 \\
s6 & 	 0.8538 & 	0.8520 & 	0.6471 & 0.6974 & 0.8715 & 0.8897  & 0.8805 	& 0.8408	& 0.9067 & 0.8645 & 0.8587 & 0.8666 \\
s7 & 	 0.9416 & 	0.9407 & 	0.8416 & 0.8723 & 0.9162 & 0.9549	& 0.9454  & 0.9374	& 0.9554 & 0.9440 & 0.9323 & 0.9457
\\\hline
\multicolumn{13}{c}{}\\
\hline
\multicolumn{13}{|c|}{(b) Cross-subject P300 detection} \\
\hline
\textbf{Subject} &
\textbf{CNN1} &  \textbf{UCNN1} & \textbf{CNN3} & \textbf{UCNN3} & \textbf{CNN-R}  &\textbf{Deep ConvNet} & \textbf{Shallow ConvNet} &  \textbf{BN$^3$} & \textbf{EEGNet} & \textbf{OCLNN} & \textbf{FCNN}  &\textbf{SepConv1D}\\
\hline
s0 & 0.7585 & 0.7361 & 0.7502 & 0.7619 & 0.7301 & 0.7370 & 0.7554 & 0.7492 & 0.7565 & 0.7546 & 0.7239 & 0.7360 \\
s1 & 0.7692 & 0.7359 & 0.7012 & 0.7126 & 0.7646 & 0.7742 & 0.7618 & 0.7467 & 0.7543 & 0.7427 & 0.7181 & 0.7600 \\
s2 & 0.8220 & 0.8344 & 0.8200 & 0.8376 & 0.8570 & 0.8275 & 0.8198 & 0.8272 & 0.8381 & 0.8369 & 0.8087 & 0.8257 \\
s3 & 0.7480 & 0.7360 & 0.6987 & 0.7471 & 0.7518 & 0.7703 & 0.7643 & 0.7585 & 0.7612 & 0.7566 & 0.7390 & 0.7515 \\
s4 & 0.7888 & 0.7687 & 0.7988 & 0.8040 & 0.8246 & 0.7970 & 0.8090 & 0.7502 & 0.8043 & 0.7529 & 0.7605 & 0.7988 \\
s5 & 0.8039 & 0.8159 & 0.8221 & 0.8269 & 0.7529 & 0.8054 & 0.7917 & 0.8028 & 0.8109 & 0.8023 & 0.7602 & 0.8023 \\
s6 & 0.7919 & 0.7724 & 0.7339 & 0.7299 & 0.8039 & 0.8210 & 0.7693 & 0.8069 & 0.8099 & 0.8001 & 0.7459 & 0.7601 \\
s7 & 0.8809 & 0.8446 & 0.8414 & 0.8411 & 0.8337 & 0.8763 & 0.8352 & 0.8491 & 0.8450 & 0.8416 & 0.8395 & 0.8158 
\\\hline
\end{tabular}
\label{tab:AUCdetail-within-als}
\end{sidewaystable}

\begin{sidewaystable}[htbp!]
\centering
\caption{Number of parameters (Param) and Floating Point Operations per Second (FLOPS) of the architectures under analysis for the four benchmark datasets. }
\begin{tabular}{|p{1.3cm}|p{2.1cm}|p{2.1cm}|p{1.5cm}|p{1.5cm}|p{1.8cm}|p{1.4cm}|p{1.4cm}|p{1.2cm}|p{1.2cm}|p{1.6cm}|}
\hline
\multicolumn{11}{|c|}{(a) ${D}_1$} \\
\hline
\textbf{Measure} &
\textbf{CNN1/UCNN1}  & \textbf{CNN3/UCNN3} & \textbf{CNN-R}  &\textbf{Deep ConvNet} & \textbf{Shallow ConvNet} &  \textbf{BN$^3$} & \textbf{EEGNet} & \textbf{OCLNN} & \textbf{FCNN}  &\textbf{SepConv1D}\\

\hline
Param  &  1,036,922 & 1,031,009   & 19,848,098 & 140,627 & 12,162 & 44,633 & 1,474 &  1,842 & 2,477 & 225\\
FLOPS &  2,073,642 &  2,061,816  & 39,683,214 & 278,976 &  24,088 & 89,304  & 2,801& 3,653 & 4,950 & 443
\\\hline
\multicolumn{11}{c}{}\\
\hline
\multicolumn{11}{|c|}{(b) ${D}_2$} \\
\hline
\textbf{Measure} &
\textbf{CNN1/UCNN1}  & \textbf{CNN3/UCNN3} & \textbf{CNN-R}  &\textbf{Deep ConvNet} & \textbf{Shallow ConvNet} &  \textbf{BN$^3$} & \textbf{EEGNet} & \textbf{OCLNN} & \textbf{FCNN}  &\textbf{SepConv1D}\\

\hline
Param  &  787,502    &  781,067     & 16,445,794  & 175,677 & 104,402 & 39,649 & 2,338 & 14,706 & 19,973 & 1,361\\
FLOPS &  1,574,802 &  1,561,932  &  32,878,606 & 349,076 &  208,568 &  79,394 & 4,529 & 29,381 & 39,942  & 2,715
\\\hline
\multicolumn{11}{c}{}\\
\hline
\multicolumn{11}{|c|}{(c) ${D}_3$} \\
\hline
\textbf{Measure} &
\textbf{CNN1/UCNN1}  & \textbf{CNN3/UCNN3} & \textbf{CNN-R}  &\textbf{Deep ConvNet} & \textbf{Shallow ConvNet} &  \textbf{BN$^3$} & \textbf{EEGNet} & \textbf{OCLNN} & \textbf{FCNN}  &\textbf{SepConv1D}\\

\hline
Param  & 1,207,502  & 1,201,067   & 21,950,818      &  177,677      & 105,362 & 47,841 & 2,434& 14,898 & 2,885 & 1,405\\
FLOPS & 2,414,802  &  2,401,932  & 43,888,654 & 29,765          & 210,488 & 95,778 & 4,721 &  353,076 & 5,766 & 2,803\\\hline
\multicolumn{11}{c}{}\\
\hline
\multicolumn{11}{|c|}{(d) ${D}_4$} \\
\hline
\textbf{Measure} &
\textbf{CNN1/UCNN1}  & \textbf{CNN3/UCNN3} & \textbf{CNN-R}  &\textbf{Deep ConvNet} & \textbf{Shallow ConvNet} &  \textbf{BN$^3$} & \textbf{EEGNet} & \textbf{OCLNN} & \textbf{FCNN}  &\textbf{SepConv1D}\\

\hline
Param  & 1,036,942  & 1,031,011   & 19,848,290 &141,877 & 15,362 & 44,673 & 1,506 & 2,290 & 3,301 & 265\\
FLOPS &  2,073,682 &  2,061,820  & 39,683,598 & 281,476 & 30,488 & 89,386 & 2,865 & 4,549 & 6,598 & 523
\\\hline
\end{tabular}
\label{tab:Complexity}
\end{sidewaystable}


\begin{thebibliography}{10}
\expandafter\ifx\csname url\endcsname\relax
  \def\url#1{\texttt{#1}}\fi
\expandafter\ifx\csname urlprefix\endcsname\relax\def\urlprefix{URL }\fi
\expandafter\ifx\csname href\endcsname\relax
  \def\href#1#2{#2} \def\path#1{#1}\fi

\bibitem{Wolpaw2000}
J.~Wolpaw, N.~Birbaumer, W.~J. Heetderks, D.~J. McFarland, P.~H. Peckham,
  aG.~Schalk, E.~Donchin, L.~A. Quatrano, C.~J. Robinson, T.~M. Vaughan,
  Brain-computer interface technology: A review of the first international
  meeting, IEEE Transactions on Rehabilitation Engineering 8~(2) (2000)
  164--173.

\bibitem{Donchin2000}
E.~Donchin, K.~M. Spencer, R.~Wijesinghe, The mental prosthesis: assessing the
  speed of a {P300}-based {Brain-Computer Interface}, {IEEE} transactions on
  rehabilitation engineering 8~(2) (2000) 174--179.

\bibitem{Niedermeyer2005}
N.~E, L.~da~Silva~F, Electroencephalography: Basic Principles, Clinical
  Applications, and Related Fields, 5th Edition, Lippincott Williams \&
  Wilkins, 2005.

\bibitem{Lawhern2018}
V.~J. Lawhern, A.~J. Solon, N.~R. Waytowich, S.~M. Gordon, C.~P. Hung, B.~J.
  Lance, {EEGNet}: A compact convolutional network for {EEG-based
  Brain-Computer Interfaces}, Journal of Neural Engineering 15~(5) (2018)
  056013.

\bibitem{Kaur2013}
M.~Kaur, P.~Ahmed, M.~Q. Rafiq, Analysis of extracting distinct functional
  components of {P300} using wavelet transform, Mathematical Models in
  Engineering and Computer Science.

\bibitem{Atum2010}
Y.~Atum, I.~Gareis, G.~Gentiletti, R.~Acevedo, L.~Rufiner, Genetic feature
  selection to optimally detect {P300 in Brain Computer Interfaces}, in:
  Engineering in Medicine and Biology Society ({EMBC}), 2010 Annual
  International Conference of the {IEEE}, IEEE, 2010, pp. 3289--3292.

\bibitem{Farwell1988}
L.~Farwell, E.~Donchin, Talking off the top of your head: toward a mental
  prosthesis utilizing event-related brain potentials, Electroencephalography
  and clinical Neurophysiology 70~(6) (1988) 510--523.

\bibitem{Alvarado2016b}
M.~Alvarado-Gonz{\'a}lez, E.~Gardu{\~n}o, E.~Bribiesca,
  O.~Y{\'a}{\~n}ez-Su{\'a}rez, V.~Medina-Ba{\~n}uelos, {P300} detection based
  on {EEG} shape features, Computational and mathematical methods in medicine
  2016 (2016) 14.

\bibitem{Krusienski2006}
D.~J. Krusienski, E.~W. Sellers, F.~Cabestaing, S.~Bayoudh, D.~J. McFarland,
  T.~M. Vaughan, J.~R. Wolpaw, A comparison of classification techniques for
  the {P300} speller, Journal of Neural Engineering 3~(4) (2006) 299.

\bibitem{Bostanov2004}
V.~Bostanov, {BCI} competition 2003-data sets {I}b and {II}b: feature
  extraction from event-related brain potentials with the continuous wavelet
  transform and the t-value scalogram, {IEEE} Transactions on Biomedical
  Engineering 51~(6) (2004) 1057--1061.

\bibitem{Kaper2004}
M.~Kaper, P.~Meinicke, U.~Grossekathoefer, T.~Lingner, H.~Ritter, B{CI}
  competition 2003-data set {IIb: Support Vector Machines for the P}300 speller
  paradigm, IEEE Transactions on Biomedical Engineering 51~(6) (2004)
  1073--1076.

\bibitem{Cechovic2013}
L.~\v{C}echovi\v{c}, M.~Hodo\v{n}, M.~Jure\v{c}ka, P300 evoked potentials data
  classification using feed forward neural network, European International
  Journal of Science and Technology 2~(2) (2013) 5.

\bibitem{Abdulhay2017}
E.~Abdulhay, R.~Oweis, A.~Mohammad, L.~Ahmad, Investigation of a wavelet-based
  neural network learning algorithm applied to {P300 based Brain-Computer
  Interface}, Biomedical Research (2017) S320--S324.

\bibitem{Woehrle2015}
H.~{Woehrle}, M.~M. {Krell}, S.~{Straube}, S.~K. {Kim}, E.~A. {Kirchner},
  F.~{Kirchner}, An adaptive spatial filter for user-independent single trial
  detection of {Event-Related Potentials}, IEEE Transactions on Biomedical
  Engineering 62~(7) (2015) 1696--1705.

\bibitem{Zeyl2016}
T.~Zeyl, E.~Yin, M.~Keightley, T.~Chau, Partially supervised {P}300 speller
  adaptation for eventual stimulus timing optimization: target confidence is
  superior to error-related potential score as an uncertain label, Journal of
  Neural Engineering 13~(2) (2016) 026008.

\bibitem{Mayaud2016}
L.~Mayaud, S.~Cabanilles, A.~V. Langhenhove, M.~Congedo, A.~Barachant,
  S.~Pouplin, S.~Filipe, L.~P{\'e}t{\'e}gnief, O.~Rochecouste, E.~Azabou,
  C.~Hugeron, M.~Lejaille, D.~Orlikowski, D.~Annane, Brain-computer interface
  for the communication of acute patients: a feasibility study and a randomized
  controlled trial comparing performance with healthy participants and a
  traditional assistive device, Brain-Computer Interfaces 3~(4) (2016)
  197--215.

\bibitem{Schirrmeister2017}
R.~T. Schirrmeister, J.~T. Springenberg, L.~D.~J. Fiederer, M.~Glasstetter,
  K.~Eggensperger, M.~Tangermann, F.~Hutter, W.~Burgard, T.~n. Ball, Deep
  learning with convolutional neural networks for {EEG} decoding and
  visualization, Human Brain Mapping 38~(11) (2017) 5391--5420.

\bibitem{Liu2018}
M.~Liu, W.~Wu, Z.~Gu, Z.~Yu, F.~Qi, Y.~Li, Deep learning based on {Batch
  Normalization} for {P300} signal detection, Neurocomputing 275 (2018)
  288--297.

\bibitem{He2016}
K.~{He}, X.~{Zhang}, S.~{Ren}, J.~{Sun}, Deep residual learning for image
  recognition, in: 2016 IEEE Conference on Computer Vision and Pattern
  Recognition (CVPR), 2016, pp. 770--778.

\bibitem{Abdel-Hamid2014}
O.~{Abdel-Hamid}, A.~{Mohamed}, H.~{Jiang}, L.~{Deng}, G.~{Penn}, D.~{Yu},
  Convolutional {Neural Networks for Speech Recognition}, IEEE/ACM Transactions
  on Audio, Speech, and Language Processing 22~(10) (2014) 1533--1545.

\bibitem{Lotte2015b}
F.~Lotte, Signal processing approaches to minimize or suppress calibration time
  in oscillatory activity-based {Brain-Computer Interfaces}, Proceedings of the
  IEEE 103~(6) (2015) 871--890.

\bibitem{Lotte2018}
F.~Lotte, L.~Bougrain, A.~Cichocki, M.~Clerc, M.~Congedo, A.~Rakotomamonjy,
  F.~Yger, A review of classification algorithms for {EEG}-based
  {Brain-Computer Interfaces: a 10 year update}, Journal of Neural Engineering
  15~(3) (2018) 031005.

\bibitem{Roy2019}
Y.~Roy, H.~Banville, I.~Albuquerque, A.~Gramfort, T.~H. Falk, J.~Faubert, Deep
  learning-based electroencephalography analysis: a systematic review, Journal
  of Neural Engineering 16~(5) (2019) 051001.

\bibitem{Cecotti2011}
H.~Cecotti, A.~Graser, Convolutional neural networks for {P300} detection with
  application to {Brain-Computer Interfaces}, {IEEE} Transactions on Pattern
  Analysis and Machine Intelligence 33~(3) (2011) 433--445.

\bibitem{Manor2015}
R.~Manor, A.~B. Geva, Convolutional neural network for multi-category rapid
  serial visual presentation {BCI}, Frontiers in Computational Neuroscience
  9~(146).

\bibitem{Ioffe2015}
S.~Ioffe, C.~Szegedy, Batch normalization: Accelerating deep network training
  by reducing internal covariate shift, in: International Conference on Machine
  Learning, 2015, pp. 448--456.

\bibitem{Srivastava2014}
N.~Srivastava, G.~Hinton, A.~Krizhevsky, I.~Sutskever, R.~Salakhutdinov,
  Dropout: A simple way to prevent neural networks from overfitting, Journal of
  Machine Learning Research 15 (2014) 1929--1958.

\bibitem{Shan2018}
H.~Shan, Y.~Liu, T.~Stefanov, A simple {C}onvolutional {N}eural {N}etwork for
  accurate {P300} detection and character spelling in {B}rain {C}omputer
  {I}nterface, in: Proceedings of the Twenty-Seventh International Joint
  Conference on Artificial Intelligence, {IJCAI-18}, International Joint
  Conferences on Artificial Intelligence Organization, 2018, pp. 1604--1610.

\bibitem{chollet2017xception}
F.~Chollet, Xception: Deep learning with depthwise separable convolutions, in:
  Proceedings of the IEEE conference on computer vision and pattern
  recognition, 2017, pp. 1251--1258.

\bibitem{howard2017mobilenets}
A.~G. Howard, M.~Zhu, B.~Chen, D.~Kalenichenko, W.~Wang, T.~Weyand,
  M.~Andreetto, H.~Adam, Mobilenets: Efficient convolutional neural networks
  for mobile vision applications, arXiv preprint arXiv:1704.04861.

\bibitem{sandler2018mobilenetv2}
M.~Sandler, A.~Howard, M.~Zhu, A.~Zhmoginov, L.-C. Chen, Mobilenetv2: Inverted
  residuals and linear bottlenecks, in: Proceedings of the IEEE conference on
  computer vision and pattern recognition, 2018, pp. 4510--4520.

\bibitem{Ledesma2010}
C.~Ledesma-Ramirez, E.~Bojorges-Valdez, O.~Y{\'a}{\~n}ez-Suarez, C.~Saavedra,
  L.~Bougrain, G.~G. Gentiletti, An open-access {P}300 speller database,
  {Fourth International Brain-Computer Interface Meeting}, poster (May 2010).

\bibitem{Blankertz2004}
B.~Blankertz, K.-R. Muller, G.~Curio, T.~M. Vaughan, G.~Schalk, J.~R. Wolpaw,
  A.~Schlogl, C.~Neuper, G.~Pfurtscheller, T.~Hinterberger, et~al., The {BCI
  competition 2003: progress and perspectives in detection and discrimination
  of EEG} single trials, IEEE Transactions on Biomedical Engineering 51~(6)
  (2004) 1044--1051.

\bibitem{Blankertz2006}
B.~Blankertz, K.-R. Muller, D.~J. Krusienski, G.~Schalk, J.~R. Wolpaw,
  A.~Schlogl, G.~Pfurtscheller, J.~R. Millan, M.~Schroder, N.~Birbaumer, The
  {BCI competition III: Validating alternative approaches to actual BCI}
  problems, IEEE transactions on neural systems and rehabilitation engineering
  14~(2) (2006) 153--159.

\bibitem{Riccio2013}
A.~Riccio, L.~Simione, F.~Schettini, A.~Pizzimenti, M.~Inghille~fri,
  M.~Olivetti~Belardinelli, D.~Mattia, F.~Cincotti, Attention and {P300-based
  BCI }performance in people with amyotrophic lateral sclerosis, Frontiers in
  Human Neuroscience 7 (2013) 732.

\bibitem{BNCI_Horizon2014}
B.~Riccio, B{NCI Horizon 2020: The Future of Brain/Neural Computer Interaction:
  H}orizon 2020, \url {http://bnci-horizon-2020.eu/database/data-sets} (2014).

\bibitem{Kingma2015}
D.~P. Kingma, J.~Ba, Adam: A method for stochastic optimization, in: 3rd.
  International Conference on Learning Representations (ICLR), San Diego, USA,
  2015.

\bibitem{Chollet2015}
F.~Chollet, et~al., Keras, \url{https://github.com/fchollet/keras} (2015).

\bibitem{Abadi2015}
M.~Abadi, {TensorFlow}: Large-scale machine learning on heterogeneous systems,
  \url {https://www.tensorflow.org/} (2015).

\bibitem{Krusienski2008}
D.~J. Krusienski, E.~W. Sellers, D.~J. McFarland, T.~M. Vaughan, J.~R. Wolpaw,
  Toward enhanced {P300} speller performance, Journal of neuroscience methods
  167~(1) (2008) 15--21.

\bibitem{Polich2006}
J.~Polich, J.~R. Criado, Neuropsychology and neuropharmacology of {P}3a and
  {P}3b, International Journal of Psychophysiology 60~(2) (2006) 172--185.

\bibitem{Lin2009}
C.-J. Lin, M.-H. Hsieh, Classification of mental task from {EEG} data using
  neural networks based on particle swarm optimization, Neurocomputing 72~(4)
  (2009) 1121--1130.

\bibitem{Wilson2019}
N.~R. Wilson, D.~Sarma, J.~D. Wander, K.~E. Weaver, J.~G. Ojemann, R.~P. Rao,
  C{ortical Topography of Error-Related High-Frequency Potentials During
  Erroneous Control in a Continuous Control Brain--Computer Interface},
  Frontiers in neuroscience 13 (2019) 502.

\bibitem{Spuler2015}
M.~Sp{\"u}ler, C.~Niethammer, Error-related potentials during continuous
  feedback: using {EEG} to detect errors of different type and severity,
  Frontiers in human neuroscience 9 (2015) 155.

\bibitem{Lecun1989}
Y.~Lecun, Generalization and network design strategies, Elsevier, 1989.

\bibitem{Goodfellow2016}
I.~Goodfellow, Y.~Bengio, A.~Courville, Deep Learning, MIT Press, 2016.

\bibitem{Nair2010}
V.~Nair, G.~E. Hinton, Rectified linear units improve restricted {B}oltzmann
  machines, in: Proceedings of the 27th International Conference on Machine
  Learning (ICML-10), 2010, pp. 807--814.

\bibitem{Clevert2015}
D.~Clevert, T.~Unterthiner, S.~Hochreiter, Fast and accurate deep network
  learning by exponential linear units {(ELUs)}, in: International Conference
  on Learning Representations (ICLR), San Juan, Puerto Rico, 2016.

\end{thebibliography}
\end{document}